\documentclass[journal]{IEEEtran}
\usepackage{blindtext}
\usepackage{graphicx}
\usepackage{amsmath}

\usepackage{subfigure}
\usepackage{epstopdf}

\usepackage{makecell}
\usepackage{booktabs}
\usepackage{multirow}
\usepackage{color}

\makeatletter
\newif\if@restonecol
\makeatother

\usepackage[linesnumbered,ruled,vlined]{algorithm2e}
\usepackage{algpseudocode}

\begin{document}
%
\title{Multi-pseudo Regularized Label for Generated Data in Person Re-Identification}

\author{Yan~Huang,
        Jingsong~Xu,
        Qiang~Wu,~\IEEEmembership{Member,~IEEE}
        Zhedong~Zheng,
        Zhaoxiang~Zhang,~\IEEEmembership{Senior~Member,~IEEE}
        and~Jian~Zhang,~\IEEEmembership{Senior~Member,~IEEE}
\thanks{}
\thanks{Yan Huang, Jingsong Xu, Qiang Wu and Jian Zhang are with the Global Big Data Technologies Centre (GBDTC), School of Electrical and Data Engineering, University of Technology Sydney, Australia. (Email: Yan.Huang-3@student.uts.edu.au, JingSong.Xu@uts.edu.au, Qiang.Wu@uts.edu.au and Jian.Zhang@uts.edu.au)}
\thanks{Zhedong Zheng is with the Centre for Artificial Intelligence (CAI), School of Software, University of Technology Sydney, Australia. (Email: Zhedong.Zheng@student.uts.edu.au)}
\thanks{Zhaoxiang Zhang is with the Research Center for Brain-Inspired Intelligence, CAS Center for Excellence in Brain Science and Intelligence Technology, Institute of Automation, Chinese Academy of Sciences, Beijing 100190, China (e-mail: zhaoxiang.zhang@ia.ac.cn).}}

\markboth{}%
{Shell \MakeLowercase{\textit{et al.}}: Bare Demo of IEEEtran.cls for Journals}

\maketitle

\begin{abstract}
Sufficient training data normally is required to train deeply learned models. However, due to the expensive manual process for labelling large number of images (\emph{i.e.}, annotation), the amount of available training data (\emph{i.e.}, real data) is always limited. To produce more data for training a deep network, Generative Adversarial Network (GAN) can be used to generate artificial sample data (\emph{i.e.}, generated data). However, the generated data usually does not have annotation labels. To solve this problem, in this paper, we propose a virtual label called Multi-pseudo Regularized Label (MpRL) and assign it to the generated data. With MpRL, the generated data will be used as the supplementary of real training data to train a deep neural network in a semi-supervised learning fashion. To build the corresponding relationship between the real data and generated data, MpRL assigns each generated data a proper virtual label which reflects the likelihood of the affiliation of the generated data to pre-defined training classes in the real data domain. Unlike the traditional label which usually is a single integral number, the virtual label proposed in this work is a set of weight-based values each individual of which is a number in (0,1] called multi-pseudo label and reflects the degree of relation between each generated data to every pre-defined class of real data.  

A comprehensive evaluation is carried out by adopting two state-of-the-art convolutional neural networks (CNNs) in our experiments to verify the effectiveness of MpRL. Experiments demonstrate that by assigning MpRL to generated data, we can further improve the person re-ID performance on five re-ID datasets, \emph{i.e.}, Market-1501, DukeMTMC-reID, CUHK03, VIPeR, and CUHK01. The proposed method obtains +6.29\%, +6.30\%, +5.58\%, +5.84\%, and +3.48\% improvements in rank-1 accuracy over a strong CNN baseline on the five datasets respectively, and outperforms state-of-the-art methods.
\end{abstract}

\begin{IEEEkeywords}
person re-identification, generated data, virtual label, semi-supervised learning.
\end{IEEEkeywords}

\IEEEpeerreviewmaketitle

\section{Introduction}
In 2014s, Generative Adversarial Network (GAN) was proposed to generate data (images) with perceptual quality \cite{goodfellow2014generative}. Since then, several improved approaches \cite{radford2015unsupervised, arjovsky2017wasserstein, gulrajani2017improved} were presented to further improve the quality of generated data. However, how to use the data is still an open question. Meanwhile, person re-identification (re-ID) is a challenging task of recognizing a person amongst different camera views. It is a typical computer vision problem that requires sufficient training data to learn a discriminative model. In the past few years, deep learning has demonstrated its performance in person re-ID by producing several state-of-the-art methods \cite{huang2017deepdiff, qian2017multi, lin2017consistent, zheng2016person, zheng2016person2, zheng2016discriminatively}. To this end, sufficient labeled training data is essential to train deeply learned models in a supervised learning fashion. Although some large datasets, \emph{e.g.}, Market-1501 \cite{zheng2015scalable}, DukeMTMC-reID \cite{ZhengZY17}, CUHK03 \cite{li2014deepreid} have been proposed. However, due to the expensive cost of data acquisition that needs to manually find corresponding labels of pedestrians who appear under different camera views, the number of images per ID in these datasets is still limited.

Using generated data to solve the problem of limited training data is a promising solution. Therefore, we attempt to use unlabeled data generated by GANs to improve the person re-ID performance further. In all existing methods by using GAN, there are two main challenging points in order to assure the better performance: 1) high quality data generated by GAN \cite{radford2015unsupervised, arjovsky2017wasserstein, gulrajani2017improved}, 2) a better strategy to use the generated data into the training model \cite{ZhengZY17}. Many works focus on the first point. This paper particularly focuses on the second point. We follow the same pipeline in \cite{ZhengZY17} that incorporates generated data with real data to train deep models in a semi-supervised learning fashion. Compared with previous attempts \cite{salimans2016improved, radford2015unsupervised} that perform semi-supervised learning in the discriminator of GANs, sufficient unlabeled generated data will directly participate in training as the supplementary of limited labeled real data in our work.

In 2017s, a related work was first proposed in \cite{ZhengZY17} that introduced a method called Label Smooth Regularization for Outliers (LSRO). This method assigns virtual labels to generated data with a uniform label distribution over all the pre-defined training classes. The uniform distribution considers weights of all the pre-defined training classes equally. More specifically, if the number of pre-defined training class is $K$, the weight of each class is equally divided into $1/K$. By doing so, LSRO shows two undesirable characteristics: 1) On the real data domain, the weights over all pre-defined training classes are identical. 2) On the generated data domain, all data share the same virtual label.

\begin{figure}[t]
    \centering
    \subfigure[]{
        \includegraphics[width=0.15\columnwidth]{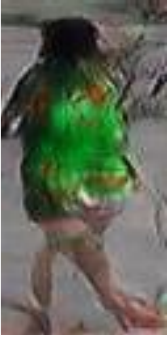}
        \label{fig:GAN_samples1}
    }
    \subfigure[]{
        \includegraphics[width=0.15\columnwidth]{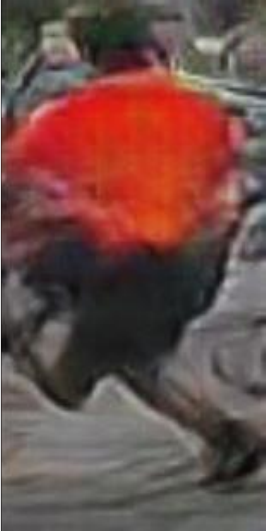}
        \label{fig:GAN_samples2}
    }
    \subfigure[Fitting generated images (a) and (b) into pre-defined training classes using markers `o' and `+', respectively.]{
        \includegraphics[width=0.58\columnwidth]{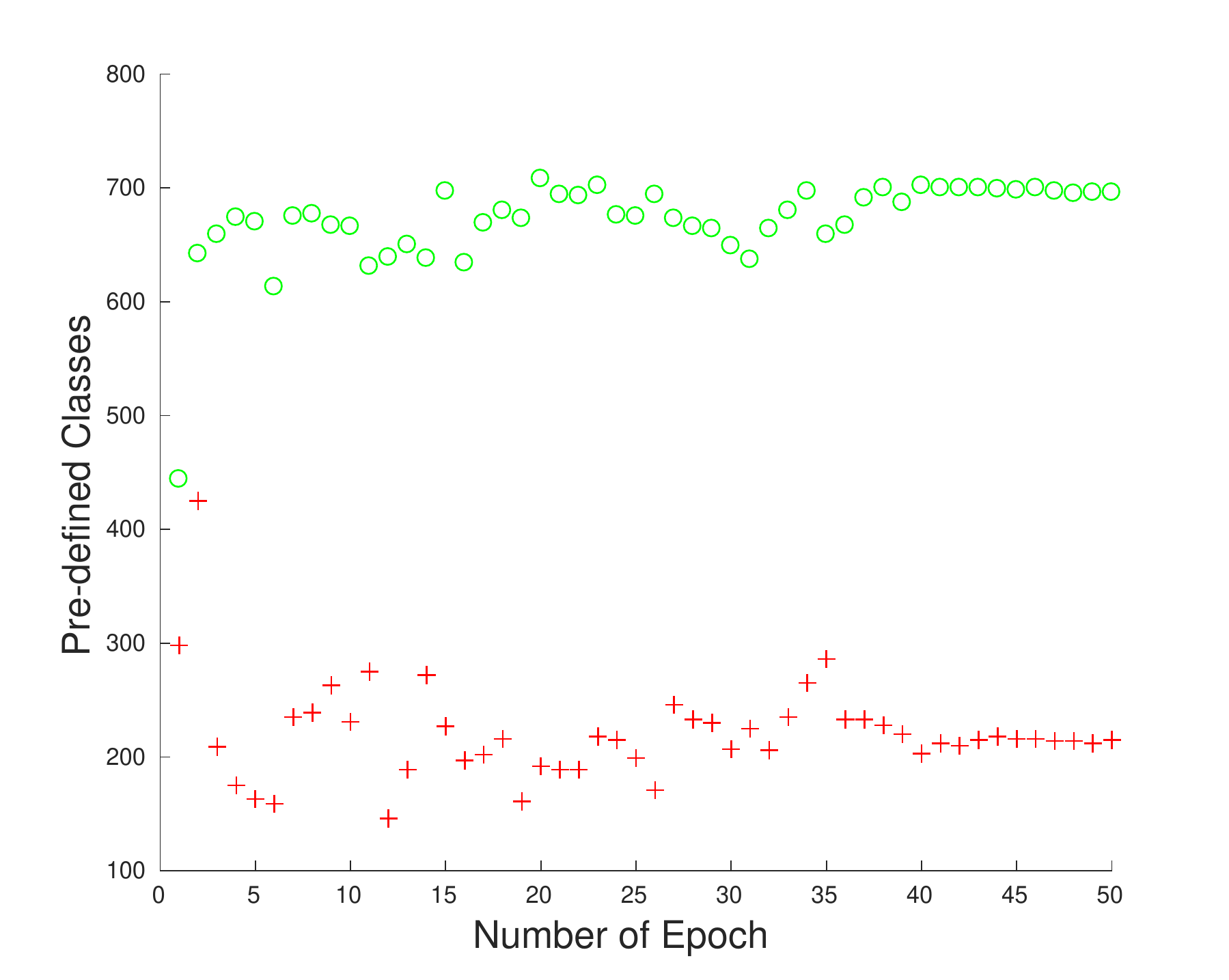}
        \label{fig:pseudo_on_2images}
    }
    \caption{Label distribution of pre-defined training classes (c) for generated images (a) and (b). Only the maximum predicted probability of pre-defined training classes is activated along with the training process (see (c)). Distinguishable label distributions can be observed between (a) and (b).}\label{fig:pseudo_over50_visualization}
\end{figure}

For the first fact, since every individual pre-defined training class of real data has the same weight, the data generated by GAN should be able to embed equal properties of all pre-defined training classes. However, during the actual GAN training process, only a random mini-batch of real data samples are used in each iteration. That is, only certain real data from some classes (not all pre-defined training classes) are used in GAN training in each iteration to generate artificial data following a continuous noise distribution \cite{goodfellow2014generative, radford2015unsupervised}. Consequently, the data distribution between the generated and real data is biased by equally utilizing the weights from all pre-defined training classes in the real data domain. \textbf{We need to assign certain type of label to generated data, which can reflect the proper weights of pre-defined training classes in GAN training on the different contributions to new data generated.} For the second fact, it may not be correct to assign the same label to certain different generated data if the generate data has the distinct visual differences. In that case, ambiguous predictions may happen in training. Figure \ref{fig:GAN_samples1} and \ref{fig:GAN_samples2} show two generated images with red and green clothes respectively. If we fit these two images into pre-defined training classes (only using the maximum predicted probability) through 50 training epochs, distinguishable label distribution can be observed in Figure \ref{fig:pseudo_on_2images}. Therefore, using the same virtual label over all the generated data is improper. \textbf{We need to dynamically assign different virtual labels to each generated data.}

Although LSRO has demonstrated its effectiveness in \cite{ZhengZY17}, the above problems still limit its effectiveness. To solve this problem, a Multi-pseudo Regularized Label (MpRL) is proposed as a virtual label assigned to generated data. Unlike LSRO, main contributions of the proposed MpRL can be summarized in three-fold:

\begin{itemize}
  \item Compared with LSRO using uniform label distribution, the proposed MpRL assigns each generated data a corresponding label which shows the likelihood of the affiliation of the generated data to all pre-defined training classes. Thus, the relationship between the generated data and pre-defined training classes can be substantially built, which makes generated data more informative when they incorporate with the real data in training.
   \item By differentiating the different generated data, MpRL can inherently mitigate of ambiguous prediction in training. Intuitively, different generated data present distinct visual differences and should have different impacts to the training. The proposed method is to embed such characteristics into the training model.
  \item Qualitative analyses are given to the proposed MpRL. Also, comprehensive quantitative evaluations are carried out to verify the performance of the proposed MpRL not only on large but also on small-scale person re-ID datasets by adapting different CNN models. In addition, we also use two groups of generated data by different GAN models to evaluate the proposed method. Such comprehensive work was not presented in \cite{ZhengZY17}.
\end{itemize}

This paper is organized as follows. We first review some related works in Section \ref{sec:related_work}. In Section \ref{sec:MpRL}, we begin to revisit the state-of-the-art virtual label used on generated data. Then the implementation details of the proposed MpRL are provided. A brief analysis is discussed to demonstrate why MpRL works better in Section \ref{sec:Why_could_MpRL_works_better}. The experiments are shown in Section \ref{sec:experimental_study}. The conclusion is in Section \ref{sec:conclusion}.

\section{Related Work}
\label{sec:related_work}
In this section, we will review existing works related to the semi-supervised learning and person re-ID.
\subsection{Semi-supervised Learning}
Semi-supervised learning is halfway between supervised and unsupervised learning, which uses both labeled and unlabeled data to perform the learning task. It has been well investigated, and dozens of methods have been proposed in the literature. In image segmentation, a small number of strongly annotated images and a large number of weakly annotated images are incorporated to perform semi-supervised learning \cite{hong2015decoupled, dai2015boxsup}. For person identification in TV series, Bauml \emph{et al}. \cite{bauml2013semi} take labeled data and unlabeled data into account and constrain them in a joint formulation. To tackle multi-label image classification, Luo \emph{et al}. \cite{luo2013manifold} make use of unlabeled data in semi-supervised learning to boost the performance. In text classification, a region embedding is learned from unlabeled data to produce additional inputs to CNN \cite{johnson2015semi}.

Since obtaining training labels is expensive, previous semi-supervised works mainly focus on how to utilize sufficient unlabeled data with accessible labeled data to boost the performance. However, if the real data is scarce or hard to obtain, these methods may useless. Therefore, in this paper, we directly use existing data to generate unlabeled data by GAN. Further, we would like to show that these generated data can help improve discriminative model learning by assigning the proposed MpRL.

Also, several methods have proposed to assign virtual labels to unlabeled data in a semi-supervised learning fashion. In \cite{Odena16a,SalimansGZCRCC16}, a new class in the discriminator is taken as the virtual label (all-in-one) assigning to all the unlabeled data produced by the generator of GAN. The all-in-one method simply regards all generated data as an extra class. Let $K$ represents the number of pre-defined training class in the real data domain, then $K+1$ is assigned to each generated data. Since these data are generated according to the distribution of real data, they tend to belong to the pre-defined training classes rather than a new one. To solve this problem, the one-hot pseudo label is proposed \cite{lee2013pseudo} that can assign a virtual label to generated data without using any extra class. The one-hot pseudo label utilizes the maximum predicted probability of the pre-defined training classes as the virtual label assigning to an unlabeled data. In training, the virtual label is dynamically assigned to the unlabeled data, so that the same data may receive a different label each time when it is fed into the network. Using the one-hot pseudo label, a generated image will be fitted into a specific pre-defined training class along with the training process, which may lead to over-fitting. To address this problem, Zheng \emph{et al}. \cite{ZhengZY17} introduce the LSRO that uses a uniform label distribution to regularize the network training for person re-ID. In this paper, the all-in-one \cite{Odena16a,SalimansGZCRCC16}, one-hot pseudo \cite{lee2013pseudo}, and LSRO \cite{ZhengZY17} will be used as our comparison experiments. Amongst them, LSRO achieves the best performance in boosting the re-ID performance. Notably, we call the pseudo \cite{lee2013pseudo} as one-hot pseudo in this paper since only one pre-defined training class with the maximum predicted probability is activated in training.

\subsection{Person Re-identification}
The person re-ID is selected to evaluate our MpRL based on two reasons. Firstly, in the past five years, there has been a tremendous increase in this research problem. It has drawn growing interest from academic researches to practical applications \cite{gong2011person}. Secondly, compared with other computer vision tasks, acquiring labeled data is expensive for person re-ID. This inspires us to leverage generated data by GAN to solve the limited training data problem. In the past few years, two branches, including traditional and deep learning methods have demonstrated their performance for person re-ID.

\textbf{In traditional methods}, the task of person re-ID can be divided into two modules: feature extraction and metric learning.
In feature extraction, Liao \emph{et al}. \cite{liao2015person} propose the local maximal occurrence feature to against viewpoint changes and handle illumination variations. Chen \emph{et al}. \cite{chen2015mirror} introduce a mirror representation to alleviate the view-specific feature distortion problem.
Zheng \emph{et al}. \cite{zheng2015scalable} present a bag-of-words descriptor that describes each person by a visual word histogram. In metric learning, Zheng \emph{et al}. \cite{zheng2011person} use a relative distance comparison method to minimize the probability of a negative person image pair that has a larger distance than a positive pair.
Liao \emph{et al}. \cite{liao2015efficient} propose logistic metric learning via an asymmetric sample weighting strategy. Li \emph{et al}. \cite{li2013learning} employ a locally-adaptive decision function that integrates traditional metric learning with a local decision rule. Yu \emph{et al}. \cite{Yu2017ICCV} learn an asymmetric metric that projects each view in an unsupervised learning fashion.

Unlike the above traditional methods that are manually designed to handle the person re-ID task. Deep learning discovers more implicit information in matching persons and achieves many state-of-the-art results.

\textbf{In deep learning methods},
to distinguish person appearance at the right spatial locations and scales, Qian \emph{et al}. \cite{qian2017multi} propose a multi-scale deep learning model to learn discriminative features. Lin \emph{et al}. \cite{lin2017consistent} introduce a consistent-aware deep learning approach which seeks the globally optimal matching. Also, deep features over the full body and body parts are captured from local context knowledge by stacking multi-scale convolutions in \cite{li2017learning}. Two-stream network \cite{geng2016deep, zheng2016discriminatively}, triplet loss network \cite{ding2015deep, cheng2016person} and quadruplet network \cite{chen2017beyond} have been designed for person re-ID.

In \cite{zheng2016person, zheng2016person2}, Zheng \emph{et al}. propose an identification (Identif) CNN. This network takes person re-ID as a multi-classification task, and a CNN embedding is learned to discriminate different identities in training. Beyond that, Zheng \emph{et al}. \cite{zheng2016discriminatively} propose a Two-stream deep neural network. A verification function that separates two input images belonging to the same or different identities is considered to improve the performance of the Identif network further. In testing, the above two networks extract CNN embeddings in the last convolutional layer to compare the similarity between two inputs using squared Euclidean distance. Both of the two CNN networks have been utilized in \cite{ZhengZY17} to investigate the improvement by adding generated data with LSRO virtual labels in training.

In this work, we adopt the Identif network \cite{zheng2016person, zheng2016person2} and the Two-stream network \cite{zheng2016discriminatively} to verify the effectiveness of the proposed MpRL. Compared with the previous related work, our MpRL achieves better performance.

\textbf{Boosting}.
In previous works, some methods have been proposed as a procedure to boost person re-ID performance further. Huang \emph{et al}. \cite{huang2016person} formulate person re-ID as a tree matching problem, and a complete bipartite graph matching is presented to refine the final matching result at the top layer of the tree. To study person re-ID with the manifold-based affinity learning, Bai \emph{et al}. \cite{bai2017scalable} introduce a manifold-preserving algorithm plunging into existing re-ID algorithms to enhance the performance. Re-ranking which exploits the relationships amongst initial ranking list in person re-ID has been studied to improve the performance \cite{zhong2017re, garcia2015person, garcia2017discriminant}. Finally, human feedback in-the-loop is required that provides an instant improvement to re-ID ranking on-the-fly \cite{ali2010interactive, wang2016human, liu2013pop}.

Unlike the above attempts, in this work, we attempt to use generated data to boost person re-ID performance on off-the-shelf CNNs by incorporating with the proposed MpRL. Although our main contribution is not to produce state-of-the-art person re-ID results. We also try to boost the performance of the Two-stream network \cite{zheng2016discriminatively} to outperform the results of several state-of-the-art methods by using our MpRL.

\section{The Proposed Multi-pseudo Regularized Label}
\label{sec:MpRL}
In this section, we first revisit the state-of-the-art virtual label LSRO \cite{ZhengZY17} for person re-ID. Then MpRL is introduced. Finally, three training strategies are given to the proposed MpRL.

\subsection{LSRO for Person Re-ID Revisit}
LSRO assumes that the generated data does not belong to any pre-defined training class and uses the uniform label distribution on each of them to address over-fitting \cite{ZhengZY17}. LSRO is inspired by label smoothing regularization (LSR) \cite{szegedy2016rethinking} which assigns less confidence on the ground-truth label and assigns small weights to other classes. Formally, giving a generated image $g$, its label distribution $q^{g}_{LSRO}(k)$ is defined as follows:
\begin{equation}
    q^{g}_{LSRO}(k)=\frac{1}{K},
    \label{equ:LSRO}
\end{equation}
where $K$ is the number of pre-defined training classes in the real data domain, $k\in \left[1,...,K\right]$ represents the $k$-th pre-defined training class. In training, the loss of LSRO to a generated image is defined as follows:
\begin{equation}
    l_{LSRO} = -\frac{1}{K}\sum_{k=1}^{K}log(p(X_{k})),
    \label{equ:LSRO_loss}
\end{equation}
where $X_{k}$ represents the output of $k$-th pre-defined training class, $p(X_{k}) \in (0,1)$ is the softmax predicted probability of $X_{k}$ belonging to the pre-defined training class $k$, defined as follows:
\begin{equation}
    p(X_{k}) = \frac{e^{X_{k}}}{\sum_{j=1}^{K} e^{X_{j}}}.
    \label{equ:softmax}
\end{equation}

\begin{itemize}
  \item In Eq.\ref{equ:LSRO_loss}, the forward loss is as follows:
\end{itemize}
\begin{equation}
\begin{aligned}
 l_{LSRO}&=-\frac{1}{K}\sum_{k=1}^{K}log(\frac{e^{X_{k}}}{\sum_{j=1}^{K}e^{X_{j}}}) \\
 &=-\frac{1}{K}\sum_{k=1}^{K}(X_{k})+log(\sum_{j=1}^{K}e^{X_{j}}).
 \label{equ:LSRO_loss_generated_forward}
\end{aligned}
\end{equation}

\begin{itemize}
  \item While, the backward gradient is as follows:
\end{itemize}
\begin{equation}
 \frac{\partial l_{LSRO}}{\partial X_{k}} = -\frac{1}{K} + \frac{e^{X_{k}}}{\sum_{j=1}^{K}e^{X_{j}}}.
 \label{equ:LSRO_loss_generated_backward}
\end{equation}

\subsection{Multi-pseudo Regularized Label}
\label{sec:Multi-pseudo_Regularization_Label_for_Outliers}
Like LSRO, we use the proposed MpRL to assign virtual labels to generated data when they are fed into the network. However, unlike LSRO, we do not set the virtual label as a uniform distribution over all pre-defined training classes (\emph{i.e.}, $1/K$). The weights over all pre-defined training classes are different in the proposed MpRL. In this way, a dictionary $\alpha$ is built to record the weights. Compared with the LSRO (see Eq.\ref{equ:LSRO}), for a generated image $g$, its label distribution is defined as follows:
\begin{equation}
    q^{g}_{MpRL}(k)=\frac{\alpha_{k}}{K},
    \label{equ:MpRL}
\end{equation}
where $\alpha_{k}$ represents the weight of $k$-th pre-defined training class in the dictionary $\alpha$. The reason why different weights are considered in the proposed MpRL will be discussed in Section \ref{sec:Using_the_Same_vs_Different_Contributions_From_Pre-defined_classes}. Our MpRL does not belong to a specifically pre-defined training class but is constituted by different weights from each of them. To obtain $\alpha_{k}$, we first formulate the set of predicted probabilities $p(X)$ of a generated image over $K$ pre-defined training classes as:
\begin{equation}
    p(X) = \left \{p(X_{k})|k\in \left[ 1,...,K \right] \right \}.
\label{equ:pred_pro_list}
\end{equation}
Then, all elements in $p(X)$ are sorted from the minimum to maximum and saved to $p_{s}(X)$:
\begin{equation}
    p_{s}(X) =\left \{p_{s}(X_{n})|n\in \left[ 1,...,K \right] \right \},
\label{equ:Psort}
\end{equation}
where $p_{s}(X_{1})==min(p(X))$ and $p_{s}(X_{K})==max(p(X))$. $\alpha_{k}$ is obtained by taking the corresponding index of $p(X_{k})$ in the set of $p_{s}(X)$:
\begin{equation}
    \alpha_{k} = \phi_{p_{s}(X)}(p(X_{k})),
\label{equ:alpha}
\end{equation}
where $\phi_{p_{s}(X)}(\cdot)$ returns the index of $p(X_{k})$ in $p_{s}(X)$. By doing so, the corresponding relationship between real data and a generated image is built by utilizing different weights obtained through the predicted probabilities over all pre-defined training classes. Combining Eq.\ref{equ:MpRL} with Eq.\ref{equ:alpha}, the proposed MpRL can assign a multiple distributed virtual label to a generated image $g$ when it is fed into the network in training:
\begin{equation}
    q^{g}_{MpRL}=\left \{\frac{\alpha_{k}}{K}|{k}\in[1,...,K]  \right \},
\label{equ:mprl2}
\end{equation}

We call our method `multi-pseudo' label because compared with the one-hot pseudo label that only the maximum predicted probability is activated, all the predicted probabilities are used in MpRL. To address over-fitting (\textit{e.g.}, after several training iterations some weights from pre-defined training classes will become larger, while others may decrease to a pretty small value), Eq.\ref{equ:mprl2} regularizes the gap between two contiguous weights to $1/K$. In this way, the proposed MpRL retains the weights from all pre-defined training classes, even though some of them may not or just producing a tiny contribution to the generated data.

Combining the generated data with real data in training, we define the cross-entropy loss of the proposed MpRL as follows:
\begin{equation}
\begin{aligned}
    l_{MpRL} = &-(1-y)log(p(X_{c})) \\
    &-y\cdot\lambda\cdot\sigma\sum_{k=1}^{K}(\frac{\alpha_{k}}{K}\cdot log(p(X_{k}))),
\label{equ:MpRL_loss}
\end{aligned}
\end{equation}
where $c$ represents the ground-truth label of a real image, $\frac{\alpha_{k}}{K}$ is defined in Eq.\ref{equ:MpRL}. $\lambda$ is the parameter for the trade-off between losses of generated and real data. If not specified, we set $\lambda$ to be 1. $\sigma$ is a normalization factor. In Eq.\ref{equ:MpRL_loss}, if we sum up weights over $K$ per-defined training classes ($\sum_{k=1}^{K}\frac{\alpha_{k}}{K}$), the total weight equals to $\frac{(1+K)\cdot K}{2}$. Therefore, to normalize weights over $K$ pre-defined training classes, $\sigma$ is set to $\frac{2}{1+K}$.

For a real image $y=0$, Eq.\ref{equ:MpRL_loss} is equivalent to softmax loss. For a generated image $y=1$, only the MpRL is used. Overall, the network has two types of losses: one for real data and the other for generated data.

\begin{itemize}
  \item In Eq.\ref{equ:MpRL_loss}, the forward loss is as follows:
\end{itemize}

For a real image, $y=0$:
\begin{equation}
\begin{aligned}
 l_{MpRL}&=-log(\frac{e^{X_{c}}}{\sum_{j=1}^{K}e^{X_{j}}}) \\
 &=-X_{c}+log(\sum_{j=1}^{K}e^{X_{j}}).
 \label{equ:MpRL_loss_real_forward}
\end{aligned}
\end{equation}

For a generated image, $y=1$:
\begin{equation}
\begin{aligned}
l_{MpRL}&=-\lambda\cdot\sigma\sum_{k=1}^{K}(\frac{\alpha_{k}}{K}\cdot log(\frac{e^{X_{k}}}{\sum_{j=1}^{K}e^{X_{j}}}))\\
&=-\lambda\cdot\sigma\sum_{k=1}^{K}(\frac{\alpha_{k}}{K}X_{k}-\frac{\alpha_{k}}{K}log(\sum_{j=1}^{K}(e^{X_{j}}))).
\label{equ:MpRL_loss_generated_forward}
\end{aligned}
\end{equation}

\begin{itemize}
  \item While, the backward gradient is as follows:
\end{itemize}

For a real image, $y=0$:
\begin{equation}
 \frac{\partial l_{MpRL}}{\partial X_{c}} = -1 + \frac{e^{X_{c}}}{\sum_{j=1}^{K}e^{X_{j}}}.
 \label{equ:MpRL_loss_real_backward}
\end{equation}

For a generated image, $y=1$:
\begin{equation}
\frac{\partial l_{MpRL}}{\partial X_{k}}=-\lambda\cdot\sigma\cdot\frac{\alpha_{k}}{K}(1-\frac{e^{X_{k}}}{\sum_{j=1}^{K}(e^{X_{j}})}).
\label{equ:MpRL_loss_generated_backward}
\end{equation}

\subsection{Training Strategy}
\label{sec:training_strategy}
To further investigate the effectiveness of the proposed MpRL, three different training strategies, including one static (constant virtual labels) and two dynamic (iteratively updated) approaches are introduced. Descriptions are as follows:
\begin{itemize}
  \item \textbf{Static MpRL (sMpRL).} The sMpRL is assigned to each generated data before training the network. We use a pre-trained Identif network (see Section \ref{sec:CNN}) to assign sMpRL. Specifically, 1) the Identif network is pre-trained on a target re-ID dataset; 2) Eq.\ref{equ:softmax} is utilized to calculate the predicted probability over $K$ pre-defined training classes for each generated data; 3) Eq.\ref{equ:mprl2} is used to assign each generated data with a sMpRL, and it remains unchanged during the whole training process. This implementation is similar to the LSRO except that we consider different weights over all pre-defined training classes instead of regarding them equally.
  \item \textbf{Dynamic MpRL-I (dMpRL-I): Dynamically Update MpRL from scratch.} During training, dMpRL-Is are dynamically assigned to each generated data using Eq.\ref{equ:mprl2}, and they will be updated iteratively to change the likelihood of the affiliation of the generated data to all pre-defined training classes. Therefore, the same generated data may receive a different dMpRL-I each time when it is fed into the network. This dynamic progress starting from the first mini-batch fed into the network until the training is completed. Notably, generated data will assign random dMpRL-Is if they are involved in the first training iteration.
  \item \textbf{Dynamic MpRL-II (dMpRL-II): Dynamically Update MpRL from the intermediate point.} We try to assign dMpRL-IIs to generated data after 20 epochs when the CNN model becomes relatively stable, and also they will be updated iteratively. That is, in Eq.\ref{equ:MpRL_loss} $y=0$, and until after 20 epochs, it is set to 1. Also, the loss is set to 0.1 and 1 for the generated and real data respectively. Therefore, under this training strategy, $\lambda$ is set to 0.1 in Eq.\ref{equ:MpRL_loss}. The detailed training strategy is shown in Algorithm \ref{alg:Dynamic MpRL from intermediate point}.
\end{itemize}
\begin{algorithm}[t]
\label{alg:Dynamic MpRL from intermediate point}
  \caption{The training strategy of the dMpRL-II: dynamically update MpRL from the intermediate point to change the likelihood of the affiliation of the generated data to all pre-defined training classes iteratively.}
  \KwIn{Real data set: $R$\;
  \qquad \quad Generated data set: $G$\;
  \qquad \quad Merged data set: $D=R\cup G$\;
  \qquad \quad Loss for the real data set: $l_{1}$\;
  \qquad \quad Loss for the generated data set: $l_{2}$.} 
  \For{number of training epochs}
  {
    Shuffle $D$ \;
        \For{number of training iterations in each epoch}
        {
          Set $l_{1}=0$, $l_{2}=0$\;
          Sample minibatch from $D\rightarrow D^{'}$\;
          Select real data $R^{'}$ from $D^{'}$\;
          Set $y=0$ in Eq.\ref{equ:MpRL_loss}\;
          Calculate loss $l_{1}$ for $R^{'}$\;
          \If{number of epochs $\geq$ 20}
          {
            Select generated data $G^{'}$ from $D^{'}$\;
            Assign MpRL to $G^{'}$ using Eq.\ref{equ:mprl2}\;
            Set $y=1$ in Eq.\ref{equ:MpRL_loss}\;
            Calculate loss $l_{2}$ for $G^{'}$\;
          }
          Calculate the final loss = $l_{1}$ + $l_{2}\times 0.1$ \;
          Backward propagation\;
          Update parameters\;
        }
  }
\textbf{final}\;
\end{algorithm}

\section{Why Multi-pseudo Regularized Label Works Better?}
\label{sec:Why_could_MpRL_works_better}
\begin{figure*}[t]
    \centering
    \includegraphics[width=2\columnwidth]{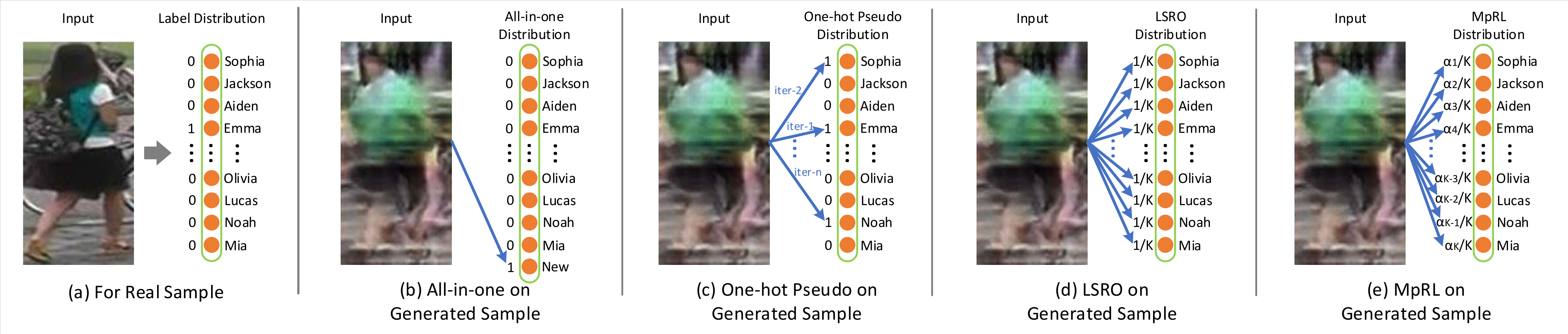}
       \caption{The label distributions of real and generated data. The ground-truth label is assigned to the real data (a). For a generated image, all-in-one (b) assigns a new label to it. One-hot pseudo (c) uses only one pre-defined training class with maximum predicted probability. LSRO (d) uses a uniform label distribution, while the proposed MpRL (e) considers different weights over all pre-defined training classes.}\label{fig:labels}
\end{figure*}

We use the all-in-one \cite{Odena16a,SalimansGZCRCC16}, one-hot pseudo \cite{lee2013pseudo}, and LSRO \cite{ZhengZY17} as our comparison experiments. Figure \ref{fig:labels}(b), (c) and (d) respectively illustrate the label distributions. Given a generated image, a new label that does not belong to any pre-defined training class is assigned to it by using the all-in-one (see Figure \ref{fig:labels}(b)). Using the one-hot pseudo, only the maximum predicted probability of pre-defined training classes is used as a virtual label (see Figure \ref{fig:labels}(c)). A uniform label distribution $1/K$ is utilized by the LSRO (see Figure \ref{fig:labels}(d)). The label distribution of MpRL is illustrated in Figure \ref{fig:labels}(e). The $\alpha = \{\alpha_{k}|k\in[1,...,K]\}$ (defined by Eq.\ref{equ:MpRL} to Eq.\ref{equ:alpha}) is used to record the different weights over all the pre-defined training classes. In this section, the differences between MpRL and the other three virtual labels will be discussed in three aspects: 1) one-hot vs. multiple label distribution, 2) the same vs. different virtual labels, and 3) the same vs. different weights from pre-defined training classes. Three qualitative discussions are given to support the MpRL, while corresponding numerical evidence will be provided in experiments (see Section \ref{sec:experimental_study}).

\subsection{One-hot vs. Multiple Label Distribution}
\label{sec:Using_One-hot_vs_Multiple_Label_Distribution}
The all-in-one and one-hot pseudo are two standard one-hot labels that assign a virtual label to each generated data outside (using a new class) and inside pre-defined training classes, respectively. Compared with the multiple label distribution that retains information from all pre-defined training classes, the one-hot distribution may produce inadequate regularization power in training which is critical to prevent the network from over-fitting. In the one-hot distribution, the network may mislead to learn a discriminative feature on an infrequent data sample or class. While using multiple distributed label, the network will discourage to be tuned towards one particular class and thus reduces the chance of over-fitting \cite{szegedy2016rethinking,ZhengZY17}. We device MpRL following the multiple label distribution. In Section \ref{sec:Comparison_with_existing_virtual_labels}, corresponding experiments demonstrate the superiority by using the multiple label distribution.

\subsection{The Same vs. Different Virtual Labels}
\label{sec:Using_the_Same_vs_Different_Virtual_Label_on_Generated_Samples}
\begin{figure}[t]
    \centering
    \includegraphics[width=0.9\columnwidth]{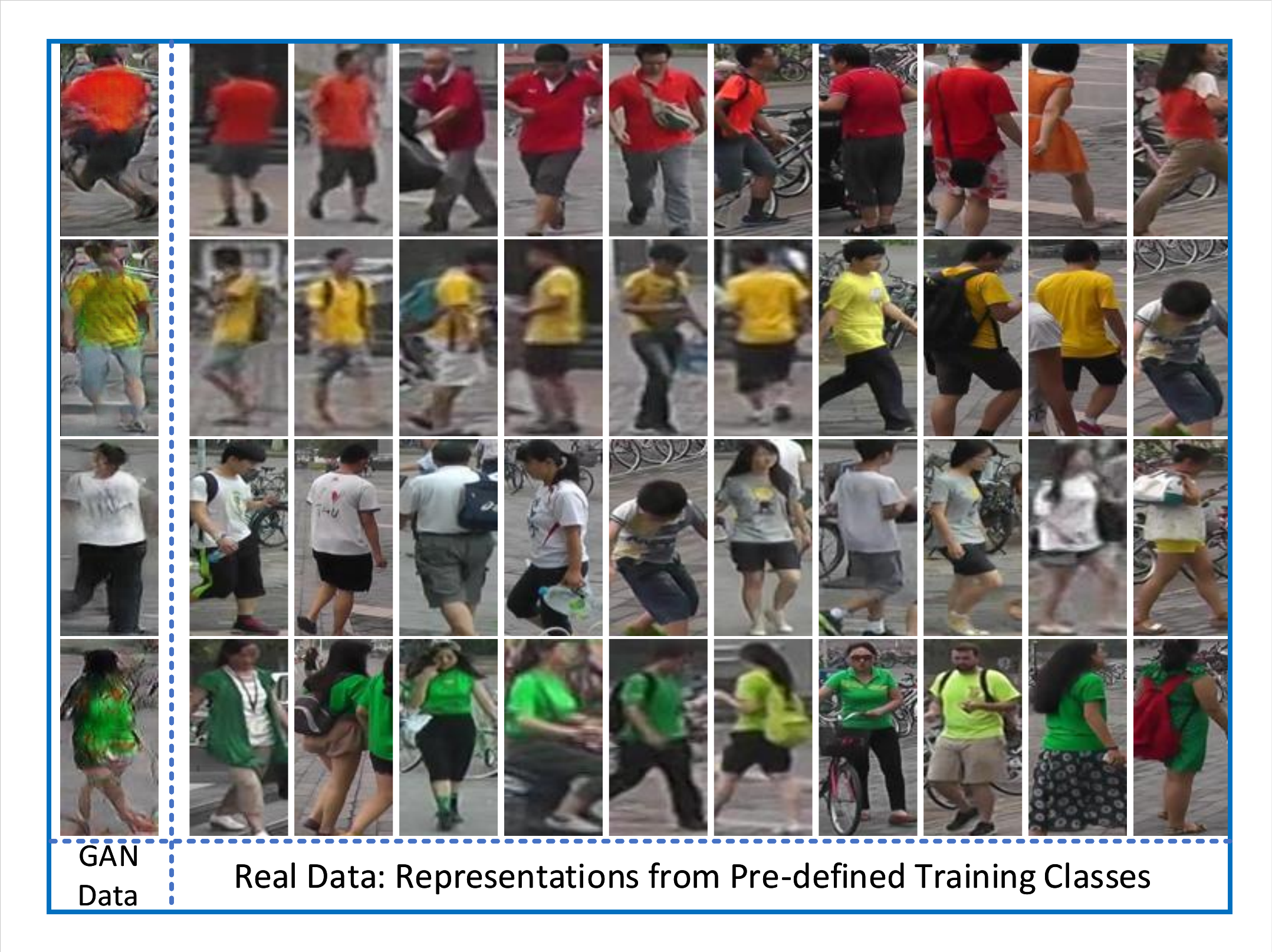}
       \caption{Examples of generated data and their corresponding representations in the real data domain. The left side shows four generated data with distinct visual differences (in red, yellow, white and green clothes). For each generated data, the right side gives ten nearest representations which represent each pre-defined training class in the real data domain. Distinguishable visual differences are shown amongst the four groups.}\label{fig:same_diff_labels}
\end{figure}
Two strategies can be used to assign virtual labels to generated data: 1) using the same virtual label over all the generated data, 2) assigning different virtual labels to different generated data. Both all-in-one and LSRO follow the first strategy, while one-hot pseudo and MpRL go with the second one. Compared with the second strategy, assigning each generated with the same label potentially leads to ambiguous predictions in training. In Figure \ref{fig:same_diff_labels}, four different generated images with distinct visual differences (in red, yellow, white and green clothes) are given to find their top ten nearest representations which represent different pre-defined training classes in the real data domain. The four groups visually show clear differences. If we still train a network by assigning the four generated images with the same virtual label, consequently, the network will mislead in identifying them. The proposed MpRL follows the second strategy that assigns each generated data with a weight-based virtual label according to different predicted probabilities in the proposed MpRL. Corresponding experiments can be found in Section \ref{sec:Comparison_with_existing_virtual_labels} to show that by assigning different virtual labels to generated data, the proposed MpRL can achieve better performance.

\subsection{The Same vs. Different Weights from Pre-defined Training Classes}
\label{sec:Using_the_Same_vs_Different_Contributions_From_Pre-defined_classes}
LSRO assumes that the weight from each pre-defined training class should be identical. Thus a generated image is assumed to have the capability to simulate the distribution of all the pre-defined training classes equally. However, this is impractical when considering the actual GAN training process, for two reasons (details can be found in \cite{goodfellow2014generative, radford2015unsupervised}). First, in each training iteration, a mini-batch of random noise is fed into a generator to simulate another mini-batch of real data. This indicates that the generation capability of the inputs is limited in a small scope, specifically, within a mini-batch of real data. Secondly, normally the input random noise obeys a continuous distribution, \emph{e.g.}, Gaussian distribution, while the distribution of real data is discrete. Consequently, complete mapping does not exist between inputs of the generator and the real data domain. Due to the above two reasons, bias exists between distributions of the output of the generator (generated data) and real data. Therefore, a generated image does not have the capability to embed equal properties of the distributions of all pre-defined training classes in the real data domain.

To address the problem of LSRO, the proposed MpRL uses different weights from pre-defined training classes (see Section \ref{sec:Multi-pseudo_Regularization_Label_for_Outliers}). In our experiment, we observe that the proposed MpRL can outperform the state-of-the-art LSRO method on three large and two small-scale person re-ID datasets (see Section \ref{sec:Comparison_with_existing_virtual_labels}).

Through the above discussion, Table \ref{tab:Properties_comparison_between_virtual_labels} summaries the properties between the proposed MpRL and other labels. Our MpRL takes the advantages of all the properties and achieves better performance than others. The numerical evidence which shows the superiority of MpRL will be presented in Section \ref{sec:experimental_study}.
\begin{table}[t]
\footnotesize
\centering
\caption{Comparison of properties amongst virtual labels, including all-in-one, one-hot pseudo, LSRO, and the proposed MpRL.}
\label{tab:Properties_comparison_between_virtual_labels}
\begin{tabular}{|c|c|c|c|}
\hline
\multicolumn{1}{|c|}{Method} & \multicolumn{1}{l|}{\begin{tabular}[c]{@{}l@{}}Label \\ Distribution\end{tabular}} & \multicolumn{1}{l|}{\begin{tabular}[c]{@{}l@{}}Label \\ Assigning\end{tabular}} & \multicolumn{1}{l|}{\begin{tabular}[c]{@{}l@{}}Weights on Pre-\\defined Classes\end{tabular}} \\ \hline
All-in-one \cite{Odena16a,SalimansGZCRCC16}  & One-hot  & Same     & --     \\ \hline
Pseudo \cite{lee2013pseudo}    & One-hot  & Different & --   \\ \hline
LSRO \cite{ZhengZY17}   & Multiple  & Same     & Same \\ \hline
MpRL (ours)  & Multiple & Different & Different  \\ \hline
\end{tabular}
\end{table}

\section{Experiments}
\label{sec:experimental_study}
In this section five person re-ID datasets are used to verify the effectiveness of the proposed MpRL, including three large-scale datasets (Market-1501 \cite{zheng2015scalable}, DukeMTMC-reID \cite{ZhengZY17}, and CUHK03 \cite{li2014deepreid}) and two small-scale datasets (VIPeR \cite{gray2008viewpoint} and CUHK01 \cite{li2012human}). We mainly evaluate the proposed MpRL using Market-1501 and VIPeR since they belong to different scales.

\subsection{Person Re-ID Datasets}
\textbf{Market-1501} is collected from six cameras in Tsinghua University. It contains 12,936 training images and 19,732 testing images. The number of identities is 751 and 750 in the training and testing sets respectively. There is an average of 17.2 images per training identity. All the pedestrians are detected by the deformable part model (DPM) \cite{felzenszwalb2010object}. Both single and multiple query settings are used.

\textbf{DukeMTMC-reID} is collected from eight cameras. The original dataset is used for cross-camera multi-target pedestrian tracking \cite{ristani2016performance}. We use the re-ID version benchmark \cite{ZhengZY17} to evaluate our method. It contains 1,404 identities in which 702 identities for training and the remaining 702 identities for testing. The total training images are 16,522. In the testing set, one query image for each identity is picked up in each camera and put the remaining images in the gallery. There are 2,228 query images and 17,661 gallery images for the 702 testing identities.

\textbf{CUHK03} is captured by six cameras on the CUHK campus. It contains 14,097 images of 1,467 identities, and each identity is observed by two disjoint camera views. There is an average of 9.6 training identity images in this set. CUHK03 contains two image settings: one is annotated by hand-drawn bounding boxes, and the other is produced by the DPM \cite{felzenszwalb2010object}. We use the detected bounding boxes and the single query setting.

\textbf{VIPeR} is a small-scale dataset that only contains 632 identities. Each identity has two images which are observed by two different camera views. There are 1,264 images in which half identities are for training and the remaining is for testing.

\textbf{CUHK01} has 971 identities, each with four images captured from two disjoint camera views. There are totally 3884 images. Two different settings can be found on this dataset: 1) 871 identities for training, and 2) 485 identities for training. We choose the latter one to verify the effectiveness of our approach since the scale of training data is much more limited than the former one. We use the multiple query setting in testing.

\subsection{Experimental Setup}
\label{sec:experimental_setup}
\subsubsection{GAN Models for Generating Data}
\label{sec:GAN}
GAN simultaneously trains two models: a generator that simulates the distribution of real data, and a discriminator that estimates the probability that a image comes from the real data set rather than the generator \cite{goodfellow2014generative}. We mainly use the DCGAN model \cite{radford2015unsupervised} and follow the same settings in \cite{ZhengZY17} for fair experimental comparisons. For the generator, 100-dim random noise is fed into a linear function to produce a tensor with size of $4\times 4\times 16$. Then, five deconvolutional functions with a kernel size of $5\times 5$ and a stride of 2 are used to enlarge the tensor. A rectified linear unit and batch normalization are used after each deconvolution. Also, one deconvolutional layer with a kernel size of $5\times 5$ and a stride of 1 are added to fine-tune the result followed by a tanh activation function. Finally, $128\times 128\times 3$ sized images can be generated. The input of the discriminator includes generated and real data. Five convolutional layers are used to classify whether the generated image is fake with a kernel size of $5\times 5$ and a stride of 2. In the end, a fully-connected layer is added to perform a binary classification.

The Tensorflow \cite{abadi2016tensorflow} and DCGAN packages are used to train the GAN model. Only data from the training set are used. All the images are resized to $128\times 128$ and randomly flipped before training. The adam stochastic optimization \cite{kingma2014adam} is used with parameters $\beta1 = 0.5, \beta2 =0.99$. The training stops after 30 and 60 epochs on large and small-scale re-ID datasets respectively. During testing, a 100-dim random vector ranged in [-1, 1] with Gaussian distribution is fed into the GAN to generate a person image. Finally, all the generated data are resized to $256\times 256$ and will be used to train CNN models with the proposed MpRL.

\begin{figure*}[t]
    \centering
    \subfigure[Market-1501.]{
        \includegraphics[width=0.35\columnwidth]{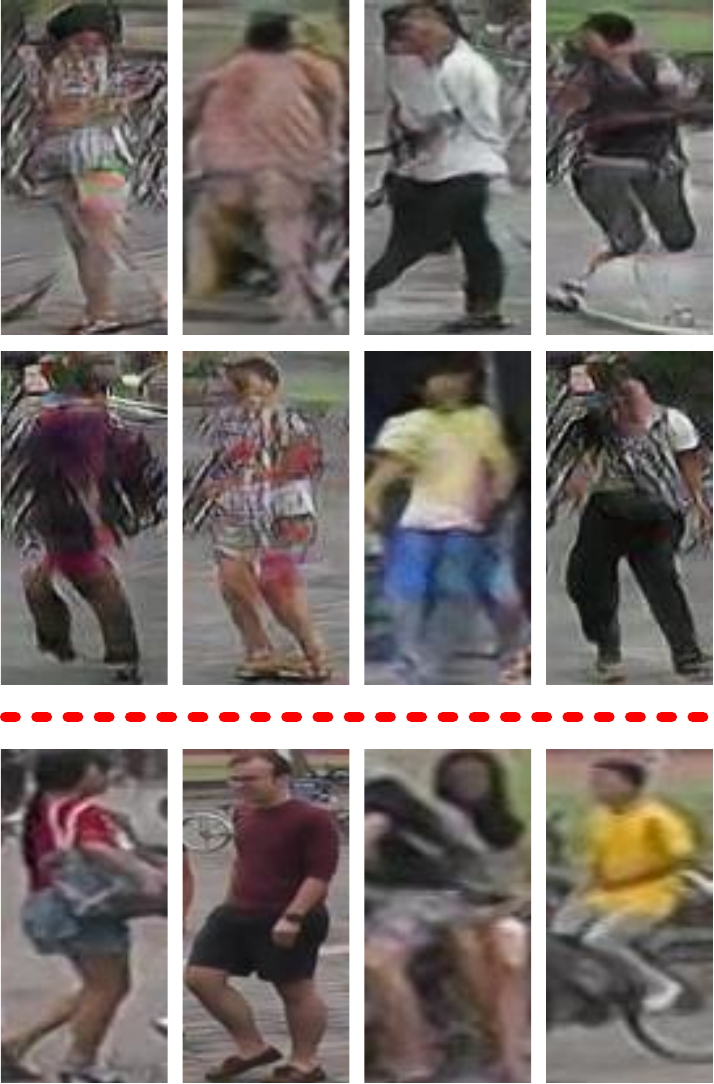}
        \label{fig:market_GAN}
    }
    \subfigure[DukeMTMC-reID.]{
        \includegraphics[width=0.35\columnwidth]{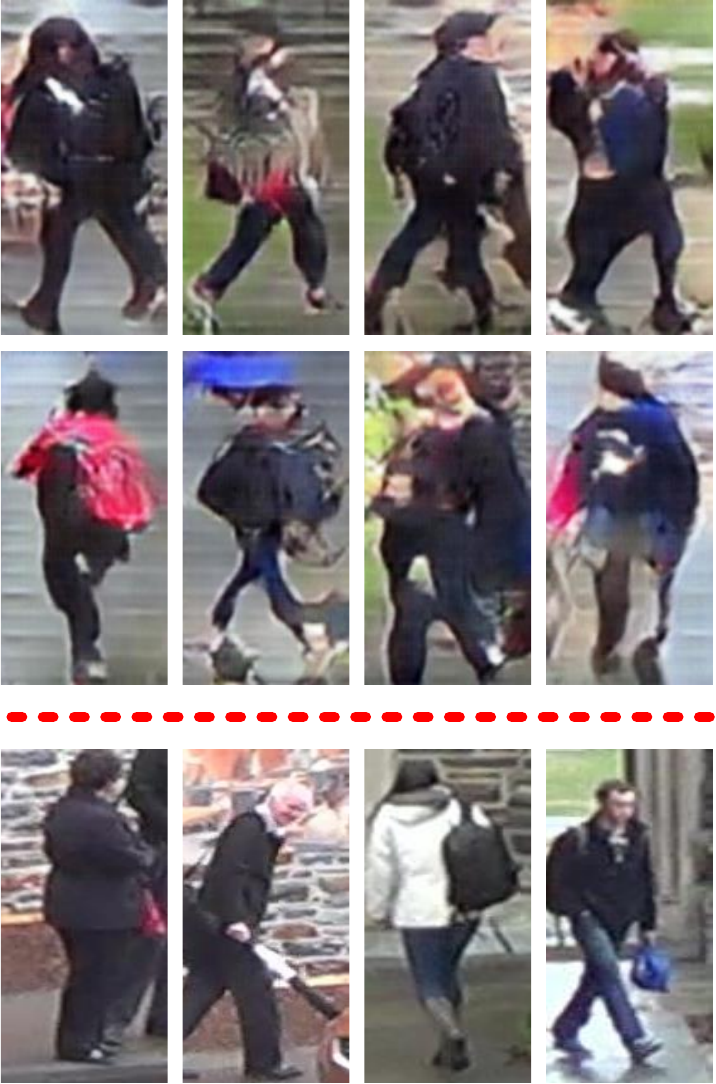}
        \label{fig:duke_GAN}
    }
    \subfigure[CUHK03.]{
        \includegraphics[width=0.35\columnwidth]{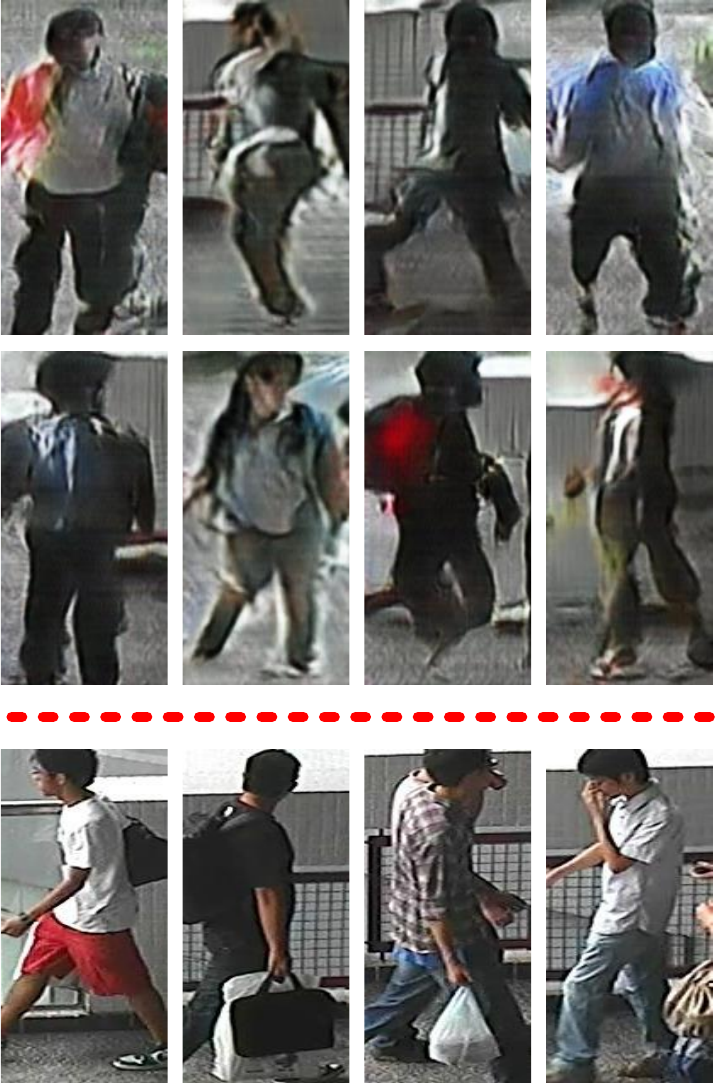}
        \label{fig:cuhk03_GAN}
    }
    \subfigure[VIPeR.]{
        \includegraphics[width=0.35\columnwidth]{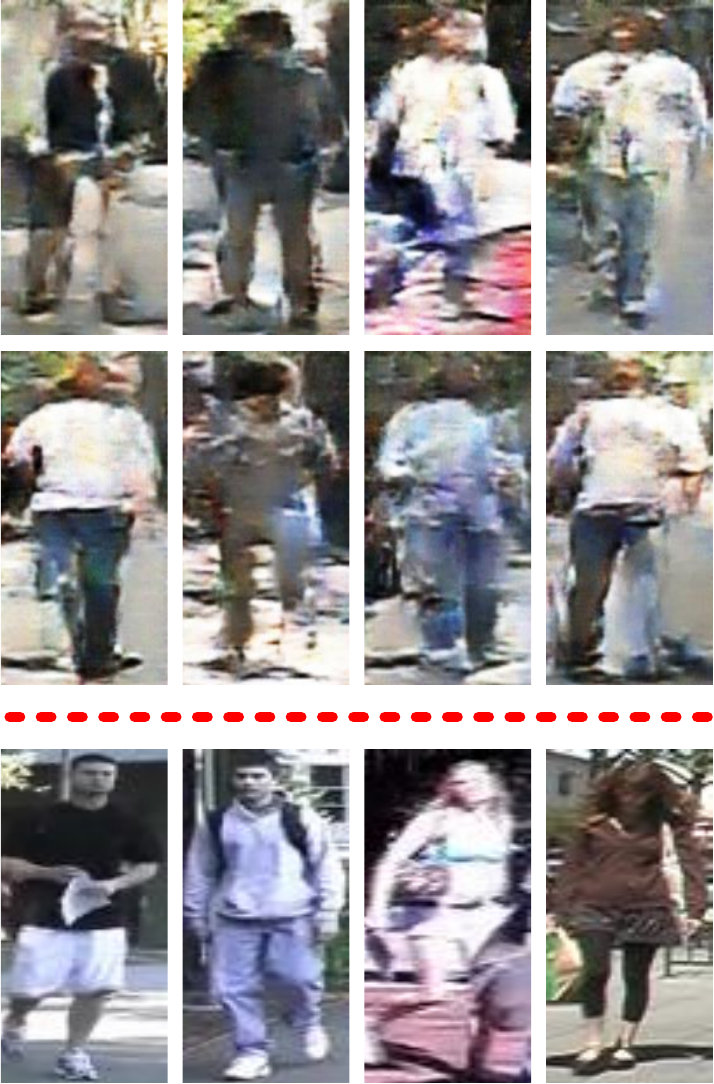}
        \label{fig:viper_GAN}
    }
    \subfigure[CUHK01.]{
        \includegraphics[width=0.35\columnwidth]{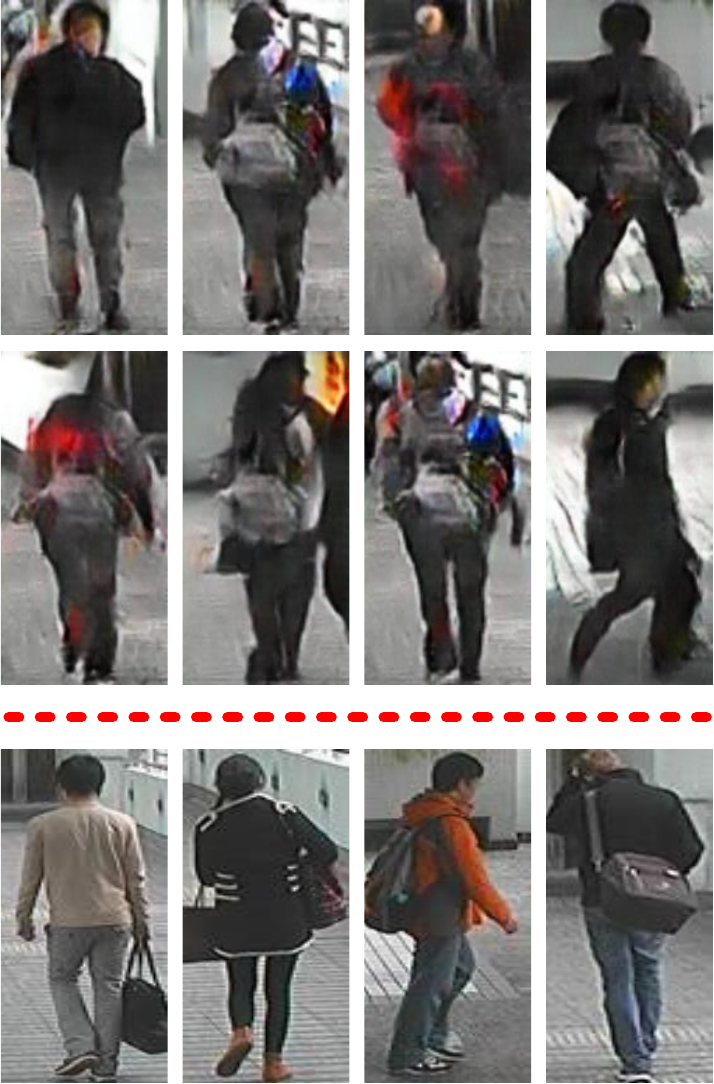}
        \label{fig:cuhk01_GAN}
    }
    \caption{Examples of generated (by DCGAN \cite{radford2015unsupervised}) and real person images. (a)-(d) show the generated person images (first two rows) and real person images (the third row) on Market-1501, DukeMTMC-reID, CUHK03, VIPeR, and CUHK01, respectively.}\label{fig:GAN_images}
\end{figure*}

Figure \ref{fig:GAN_images} illustrates the generated and real data on the five different re-ID datasets. Although the generated data can be easily recognized as fake by human, they remain effective in improving the performance by adding the proposed MpRL as virtual labels in experiments.

\subsubsection{CNNs for Evaluation}
\label{sec:CNN}
We adopt two CNNs to evaluate the proposed MpRL. These two networks have been used to evaluate the performance of the all-in-one, one-hot pseudo, and LSRO labels in \cite{ZhengZY17}. The first is an Identif network \cite{zheng2016person, zheng2016person2} that takes person re-ID as a multi-classification task according to the number of pre-defined training classes in the real data domain. We use the Identif network as a baseline when only the real data is used. Furthermore, to compare the performance of different virtual labels, generated images are incorporated into real images as inputs. The second one is a Two-stream network \cite{zheng2016discriminatively} that combines the Identif network with a verification function to train the network. Given two input images, the verification function will classify them into two classes (belong to the same or different identities). We use this Two-stream network to achieve better results by adding generated data in training. In our experiment, both Identif and Two-stream networks use the pre-trained resnet-50 \cite{he2016deep} as a basic component. We change the last fully-connected layer to have $K$ neurons to predict $K$ classes, where $K$ is the number of pre-defined training classes. Since we do not need to add extra classes on generated data by using the proposed MpRL, the last fully-connected layer remains $K$ neurons in training.

\begin{figure}[t]
    \centering
    \includegraphics[width=0.9\columnwidth]{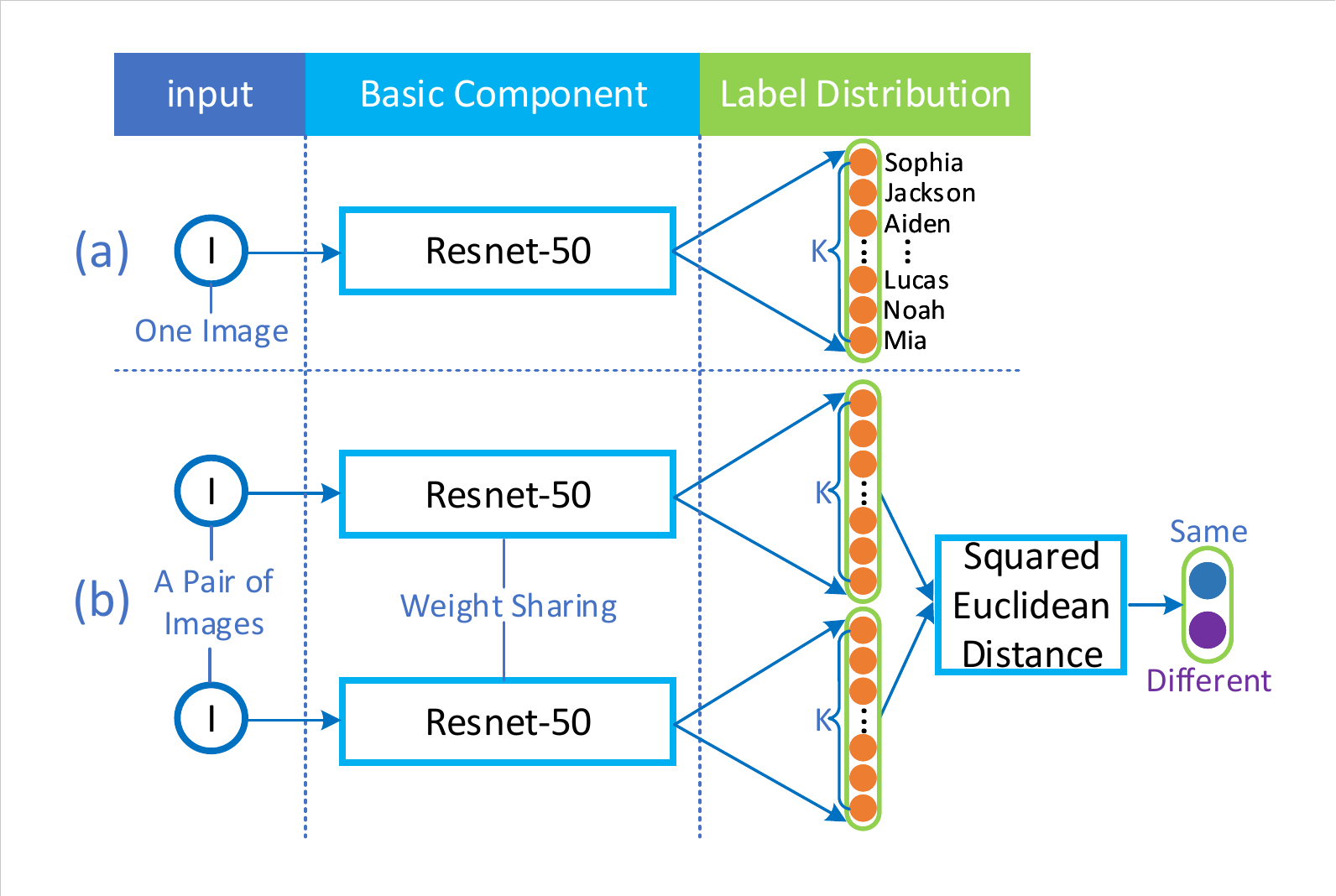}
       \caption{(a) is the Identif network presented in \cite{zheng2016person, zheng2016person2}, (b) is the Two-stream network introduced in \cite{zheng2016discriminatively}. Both networks use resnet-50 as a basic component of CNN.}\label{fig:CNN_model}
\end{figure}

Figure \ref{fig:CNN_model}(a) and Figure \ref{fig:CNN_model}(b) respectively show the Identif and Two-stream networks. MpRLs are assigned to generated data when they are fed into the network. In the Two-stream network, squared Euclidean distance is used as a similarity measure between two outputs of the $K$ neurons, and parameters are shared between the two resnet-50 components. Since generated images are unlabeled data that do not belong to any classes, only real images participate in the verification function.

The Matconvnet \cite{vedaldi2015matconvnet} package is used to implement the Identif network and the Two-stream network. All the images are resized to $256\times 256$ before being randomly cropped into $224\times 224$ with random horizontal flipping. A dropout layer is inserted before the final convolutional layer of the resnet-50. The dropout rate is set to 0.75 for Market-1501 and DukeMTMC-reID, and 0.5 for CUHK03, VIPeR, and CUHK01. We modify the fully-connected layer of resnet-50 to have 751, 702, 1,367, 316 and 485 neurons for Market-1501, DukeMTMC-reID, CUHK03, VIPeR, and CUHK01 respectively. For the verification function in the Two-stream network, a dropout layer with a rate of 0.9 is adopted after the similarity measure. Stochastic gradient descent is used on both networks with momentum 0.9. The learning rate is set to 0.1 and decay to 0.01 after 40 epochs, and we stop training after the 50-th and 60-th epochs on the Identif network and Two-stream network, respectively. For the Identif network, the batchsize is set to 64. For the Two-stream network, the batchsize is set to 32 and 48 on large and small-scale re-ID datasets respectively. During testing, for both networks, a 2,048-dim CNN embedding in the last convolutional layer of the resnet-50 is extracted. The similarity between two images is calculated by a squared Euclidean distance before ranking. Naturally, the small-scale dataset cannot train a network from the scratch. In order to build certain initial network parameters, we first use the three large scale re-ID datasets to pre-train two evaluation CNN models which we use in our experiments (\textit{i.e.}, the Identif network and the Two-stream network). Then, small datasets VIPeR and CUHK01 along with the generated data (based on the proposed method in this paper) are to fine-tune the network.

\subsection{The CNN Performance}
\label{sec:The_CNN_performance}
Using the experimental setup in Section \ref{sec:experimental_setup}, we train the Identif and Two-stream networks on Market-1501, DukeMTMC-reID, CUHK03, VIPeR and CUHK01, respectively. Table \ref{tab:baseline} shows the experimental results using the real data only. With the Identif (Two-stream) network, we obtain the rank-1 accuracy 74.08\% (81.83\%), 61.94\% (72.62\%), 63.10\% (81.88\%), 40.76\% (51.84\%), and 65.33\% (77.78\%) on Market-1501, DukeMTMC-reID, CUHK03, VIPeR, and CUHK01, respectively. The result shown in Table \ref{tab:baseline} is a baseline, and our goal is to improve the performance of the two networks by using the proposed MpRL with generated data in training.

\begin{table}[t]
\footnotesize
  \centering
  \caption{Performance of the Identif and Two-stream networks. Only the real images are used. Rank-1 accuracy and mAP are listed.}
    \begin{tabular}{c|c|c|c}
    \toprule[1.1pt]\hline
    Dataset & CNN & mAP   & rank-1 \\ \hline
    \multirow{2}[2]{*}{Market-1501} & Identif \cite{zheng2016person, zheng2016person2} & 52.68\% & 74.08\% \\
          & Two-stream \cite{zheng2016discriminatively} & 64.09\% & 81.83\% \\
    \hline
    \multirow{2}[2]{*}{DukeMTMC-reID} & Identif \cite{zheng2016person, zheng2016person2} & 42.20\% & 61.94\% \\
          & Two-stream \cite{zheng2016discriminatively} & 51.04\% & 72.62\% \\
    \hline
    \multirow{2}[2]{*}{CUHK03} & Identif \cite{zheng2016person, zheng2016person2} &  68.36\%  & 63.10\%  \\
          & Two-stream \cite{zheng2016discriminatively} &   85.20\%    &  81.88\% \\
    \hline
    \multirow{2}[2]{*}{VIPeR} & Identif \cite{zheng2016person, zheng2016person2} &   46.38\%    &  40.76\% \\
          & Two-stream \cite{zheng2016discriminatively} &   59.38\%    &  51.84\% \\
    \hline
    \multirow{2}[2]{*}{CUHK01} & Identif \cite{zheng2016person, zheng2016person2} &   63.60\%    &  65.33\% \\
          & Two-stream \cite{zheng2016discriminatively} &   76.38\%    &  77.78\% \\
    \hline\bottomrule[1.1pt]
    \end{tabular}%
  \label{tab:baseline}%
\end{table}%

\subsection{Generated Data Improve the Performance of The Identif Network}
\label{sec:The_generated_samples_improve_the_Identif network's_performance}
We first give the result of the Identif network to evaluate our MpRL. Since the performance of the Two-stream network is higher, it will be used to compare with some state-of-the-art methods with the proposed MpRL in Section \ref{sec:Comparison_with_the_state-of-the-art_methods}. Table \ref{tab:virtual_labels_3datasets} shows that when we add 24,000 GAN generated images to train the Identif network on three large-scale datasets, our dMpRL-II significantly improves the re-ID performance on the strong baseline of Market-1501. The improvements are +5.91\% (from 52.68\% to 58.59\%) and +6.29\% (from 74.08\% to 80.37\%) in mAP and rank-1 accuracy, respectively. For DukeMTMC-reID, +6.38\% (from 42.20\% to 48.58\%) and +6.30\% (from 61.94\% to 68.24\%) improvements are obtained in mAP and rank-1 accuracy, respectively. For CUHK03, the improvements are +5.12\% (from 68.36\% to 73.48\%) and +5.58\% (from 63.10\% to 68.68\%) in mAP and rank-1 accuracy, respectively. We also test the effectiveness of our proposed method on two small-scale datasets, including VIPeR and CUHK01. +5.87\% (mAP) and +5.84\% (rank-1) improvements can be observed on VIPeR by adding 1,200 generated images in training. Meanwhile, +2.77\% (mAP) and +3.48\% (rank-1) improvements can be observed on CUHK01 by adding 4,000 generated images in training. The above results indicate the proposed MpRL can effectively yield improvements over the baseline performance on both large and small-scale re-ID datasets.

\begin{table}[t]
  \centering
  \caption{Comparison between LSRO and dMpRL-II on five datasets. Identif network is used by adding 24,000, 1,200, and 4,000 generated images on the three large re-ID datasets, VIPeR, and CUHK01, respectively. We show the improvements in the \textit{italic} and \textbf{bold} font by using LSRO and the proposed MpRL, respectively.}
    \begin{tabular}{c|c|ccc}
    \toprule[1.1pt]\hline
    Dataset & Method & mAP & rank-1 \\ \hline
    \multirow{5}{*}{Market-1501} & baseline & 52.68\% & 74.08\% \\ \cline{2-4}
     & LSRO \cite{ZhengZY17} & 56.33\% & 78.21\% \\
     & Improvement & \textit{+3.65\%} & \textit{+4.14\%} \\ \cline{2-4}
     & dMpRL-II & 58.59\% & 80.37\% \\
     & Improvement & \textbf{+5.91\%} & \textbf{+6.29\%} \\ \hline
    \multirow{5}{*}{DukeMTMC-reID} & baseline & 42.20\% & 61.94\% \\ \cline{2-4}
     & LSRO \cite{ZhengZY17} & 46.66\% & 66.92\% \\
     & Improvement & \textit{+4.46\%} & \textit{+4.98\%} \\ \cline{2-4}
     & dMpRL-II & 48.58\% & 68.24\% \\
     & Improvement & \textbf{+6.38\%} & \textbf{+6.30\%} \\ \hline
    \multirow{5}{*}{CUHK03} & baseline & 68.36\% & 63.10\% \\ \cline{2-4}
     & LSRO \cite{ZhengZY17} & 71.60\% & 66.30\% \\
     & Improvement & \textit{+3.24\%} & \textit{+3.20\%} \\ \cline{2-4}
     & dMpRL-II & 73.48\% & 68.68\% \\
     & Improvement & \textbf{+5.12\%} & \textbf{+5.58\%} \\ \hline
    \multirow{5}{*}{VIPeR} & baseline & 46.38\% & 40.76\% \\ \cline{2-4}
     & LSRO \cite{ZhengZY17} & 49.94\% & 43.57\% \\
     & Improvement & \textit{+3.56\%} & \textit{+2.81\%} \\ \cline{2-4}
     & dMpRL-II & 52.25\% & 46.60\% \\
     & Improvement & \textbf{+5.87\%} & \textbf{+5.84\%} \\ \hline
    \multirow{5}{*}{CUHK01} & baseline & 63.60\% & 65.33\% \\ \cline{2-4}
     & LSRO \cite{ZhengZY17} & 64.47\% & 66.98\% \\
     & Improvement & \textit{+0.87\%} & \textit{+1.65\%} \\ \cline{2-4}
     & dMpRL-II & 66.37\% & 68.81\% \\
     & Improvement & \textbf{+2.77\%} & \textbf{+3.48\%} \\
    \hline\bottomrule[1.1pt]
    \end{tabular}%
  \label{tab:virtual_labels_3datasets}%
\end{table}%

\subsection{Comparison with Different Implementations of MpRL}
\label{sec:Comparison_with_different_MpRL_settings}
Three implementations are used in our experiments to demonstrate the effectiveness of the proposed MpRL (see Section \ref{sec:Multi-pseudo_Regularization_Label_for_Outliers}). We conduct this experiment using the Identif network. Table \ref{tab:market_iden_MpRL} gives the comparisons on Market-1501. We observe that by dynamically updating the likelihood of the affiliation of the generated data to all pre-defined training classes in training, dMpRL-I (+4.74\% and +4.87\% improvements in mAP and rank-1 accuracy respectively) and dMpRL-II (+5.91\% and +6.29\% improvements in mAP and rank-1 accuracy respectively) achieve better improvements compared with the sMpRL (+3.08\% and +4.77\% improvements in mAP and rank-1 accuracy respectively). This is because each generated data will receive a proper MpRL along with the discriminative power of the CNN getting better in training. Also, compared with dMpRL-I, dMpRL-II achieves the best improvement when the network becomes relatively stable after 20 training epochs.
\begin{table*}[t]
  \centering
  \caption{Comparison of all-in-one, one-hot pseudo, LSRO, and MpRLs under different numbers of generated data on Market-1501 by using the Identif network. The best improvement of different methods is highlighted in \textbf{bold}. Rank-1 accuracy and mAP are shown.}
    \begin{tabular}{c|c|c|c|c|c|c|c|c|c|c|c|c}
    \toprule[1.1pt]\hline
    \multirow{2}{*}{\#GAN Img} & \multicolumn{2}{c|}{All-in-one \cite{Odena16a,SalimansGZCRCC16}} & \multicolumn{2}{c|}{One-hot Pseudo \cite{lee2013pseudo}} & \multicolumn{2}{c|}{LSRO \cite{ZhengZY17}} & \multicolumn{2}{c|}{sMpRL} & \multicolumn{2}{c|}{dMpRL-I} & \multicolumn{2}{c}{dMpRL-II} \\
\cline{2-13}          & \multicolumn{1}{c|}{mAP} & rank-1 & mAP   & rank-1 & mAP   & rank-1 & mAP   & rank-1 & mAP   & rank-1 & mAP   & rank-1 \\
    \hline
    0 (base) & 52.68\% & 74.08\% & 52.68\% & 74.08\% & 52.68\% & 74.08\% & 52.68\% & 74.08\% & 52.68\% & 74.08\% & 52.68\% & 74.08\% \\
    \hline
    12000 & 55.68\% & 76.96\% & 55.69\% & 76.52\% & 55.22\% & 77.17\% & 55.27\% & 77.73\% & 55.84\% & 77.88\% & 58.14\% & 79.22\% \\
    18000 & 55.59\% & \textbf{77.40\%} & 56.04\% & 77.95\% & 55.28\% & 76.96\% & 55.05\% & 77.73\% & 56.21\% & 78.36\% & 58.31\% & 79.81\% \\
    24000 & 56.07\% & 77.21\% & \textbf{56.90\%} & 77.62\% & \textbf{56.33\%} & \textbf{78.21\%} & 55.59\% & \textbf{78.85\%} & 56.10\% & 77.79\% & \textbf{58.59\%} & \textbf{80.37\%} \\
    30000 & \textbf{56.19\%} & 77.17\% & 56.54\% & \textbf{77.95\%} & 55.40\% & 77.46\% & \textbf{55.76\%} & 77.82\% & 57.15\% & 78.65\% & 57.69\% & 79.16\% \\
    36000 & 55.24\% & 75.92\% & 56.38\% & 77.42\% & 55.82\% & 77.91\% & 55.45\% & 78.32\% & \textbf{57.42\%} & \textbf{78.95\%} & 57.61\% & 79.90\% \\
    48000 & 53.98\% & 75.16\% & 55.86\% & 76.72\% & 54.87\% & 76.90\% & 55.02\% & 77.45\% & 56.01\% & 77.57\% & 57.03\% & 78.73\% \\
    \hline
    improvement & +3.51\% & +3.32\% & +4.22\% & +3.87\% & +3.65\% & +4.13\% & +3.08\% & +4.77\% & +4.74\% & +4.87\% & \textbf{\textbf{+5.91\%}} & \textbf{\textbf{+6.29\%}} \\
    \hline\bottomrule[1.1pt]
    \end{tabular}%
  \label{tab:market_iden_MpRL}%
\end{table*}

\subsection{Comparison with Existing Virtual Labels}
\label{sec:Comparison_with_existing_virtual_labels}
To further evaluate the proposed MpRL, we compare it with other three competitive virtual labels: all-in-one, one-hot pseudo, and LSRO. Amongst them, LSRO \cite{ZhengZY17} is the state-of-the-art method using generated data for person re-ID. Table \ref{tab:market_iden_MpRL} provides the comparison results. We add a different number of generated data in training to show the improvement. By adding 30,000 and 18,000 generated images, the all-in-one achieves the best improvements in mAP (+3.51\%) and rank-1 accuracy (+3.32\%), respectively. The one-hot pseudo achieves +4.22\% (mAP) and +3.87\% (rank-1) improvements when 24,000 and 30,000 generated images are respectively added. Compared with them, LSRO obtains a better rank-1 accuracy improvement (+4.13\%) when adding 24,000 generated images. However, the improvement of mAP (+3.65\%) is slightly less than the one-hot pseudo. In this experiment, we use the same generated data over all the methods; the improvements are on par with that reported in \cite{ZhengZY17}. Although the improvement of mAP (+3.08\%) is less than other virtual labels by using sMpRL, we obtain better rank-1 accuracy improvements under all the implementations of the proposed MpRL (+4.77\%, +4.87\%, and +6.29\%, respectively). dMpRL-I and dMpRL-II also outperform other methods in mAP by +4.74\% and +5.91\% respectively. By adding 24,000 generated images, dMpRL-II improves the mAP and rank-1 accuracy of the Identif network from 52.68\% and 74.08\% to 58.59\% and 80.37\%, respectively. Our method outperforms the previous state-of-the-art method LSRO to a certain degree (mAP: +3.65\%$\rightarrow$ +5.91\%, rank-1 accuracy: +4.13\%$\rightarrow$ +6.29\%). It can be observed that when 12,000 generated images are used, there is limited regularization capability to improve the re-ID performance over all the virtual labels. Meanwhile, if too many generated images are added in training, \textit{e.g.}, 48,000, the performance is dropped since the network tends to converge towards the generated data instead of real data. To balance the number of generated data in training, we empirically set it to 24,000 over the three large-scale datasets we used.

In Table \ref{tab:market_iden_MpRL}, it is clear to see that the multiple label distribution (LSRO and MpRL) can always outperform the one-hot label distribution (all-in-one and one-hot pseudo) in the rank-1 accuracy. The reason can be found in Section \ref{sec:Using_One-hot_vs_Multiple_Label_Distribution}. Besides, we also find that compared with the way using the same label, assigning different labels to generated data can achieve better results in both multiple (MpRL vs. LSRO) and one-hot (one-hot pseudo vs. all-in-one) label distribution. The reason can be found in Section \ref{sec:Using_the_Same_vs_Different_Virtual_Label_on_Generated_Samples}.

To further investigate the performance of the proposed MpRL, we also evaluate it on two small-scale re-ID datasets. Table \ref{tab:viper_iden_MpRL} lists the result on VIPeR. Our dMpRL-II improves the mAP and rank-1 accuracy on this dataset by +5.87\% and +5.84\% respectively when adding 1,200 generated images in training, and outperforms the LSRO method. Since VIPeR is a small dataset (only 632 images for training), adding too many generated images, \textit{e.g.}, 12,000 leads to inferior results. Therefore, we set the number of generated data to approximate double that of the number of real data on small datasets. Specifically, we use 1,200 and 4,000 generated images for VIPeR and CUHK01 respectively. We mainly report the result on VIPeR by changing the number of generated data. The results of CUHK01 can be found in Table \ref{tab:virtual_labels_3datasets} and \ref{tab:state_of_the_art}.

Using the Identif network, Table \ref{tab:virtual_labels_3datasets} shows comparison results between our dMpRL-II and LSRO on three large-scale datasets by adding 24,000 generated images. Also, two small-scale datasets are used to evaluate the proposed method by adding 1,200 and 4,000 images respectively. By using different weights from pre-defined training classes, dMpRL-II can always outperform previous state-of-the-art virtual label LSRO over the five datasets. The reason can be found in Section \ref{sec:Using_the_Same_vs_Different_Contributions_From_Pre-defined_classes}.

\begin{table}[t]
  \centering
  \caption{Comparison of LSRO and the proposed dMpRL-II under different numbers of generated data on VIPeR with the Identif network. The best improvement of different methods is highlighted in \textbf{bold}. Rank-1 accuracy and mAP are listed.}
    \begin{tabular}{c|c|c|c|c}
    \toprule[1.1pt]\hline
    \multirow{2}{*}{\#GAN Img} & \multicolumn{2}{c|}{LSRO \cite{ZhengZY17}} & \multicolumn{2}{c}{dMpRL-II} \\
\cline{2-5}          
	& mAP & rank-1 & mAP  & rank-1  \\
    \hline
    0 (base) & 46.38\% & 40.76\% & 46.38\% & 40.76\% \\
    \hline
    600  & 48.98\% & 42.80\% & 48.59\% & 42.61\% \\
    1200 & \textbf{49.94\%} & \textbf{43.57\%} & \textbf{52.25\%} & \textbf{46.60\%} \\
    1800 & 49.41\% & 43.39\% & 50.51\% & 44.24\% \\
    2400 & 45.95\% & 40.65\% & 49.36\% & 43.77\% \\
    12000& 43.34\% & 37.12\% & 44.25\% & 37.66\% \\
    \hline
    improvement & +3.56\% & +2.81\% & \textbf{+5.87\%} & \textbf{+5.84\%}  \\
    \hline\bottomrule[1.1pt]
    \end{tabular}%
  \label{tab:viper_iden_MpRL}%
\end{table}

\begin{figure*}[t]
    \centering
    \subfigure[Market-1501.]{
        \includegraphics[width=0.6\columnwidth]{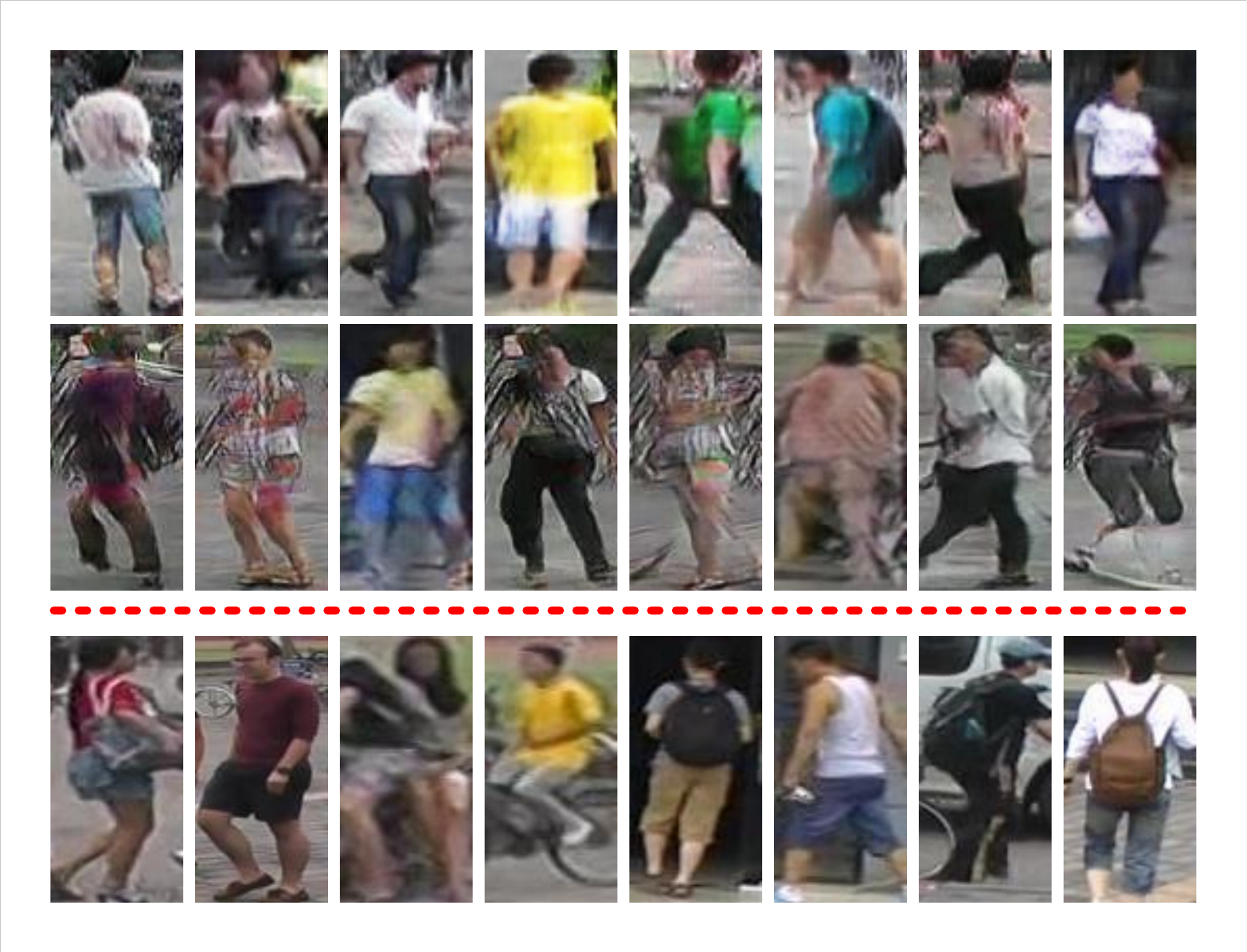}
        \label{fig:market_wgangp}
    }
    \subfigure[DukeMTMC-reID.]{
        \includegraphics[width=0.6\columnwidth]{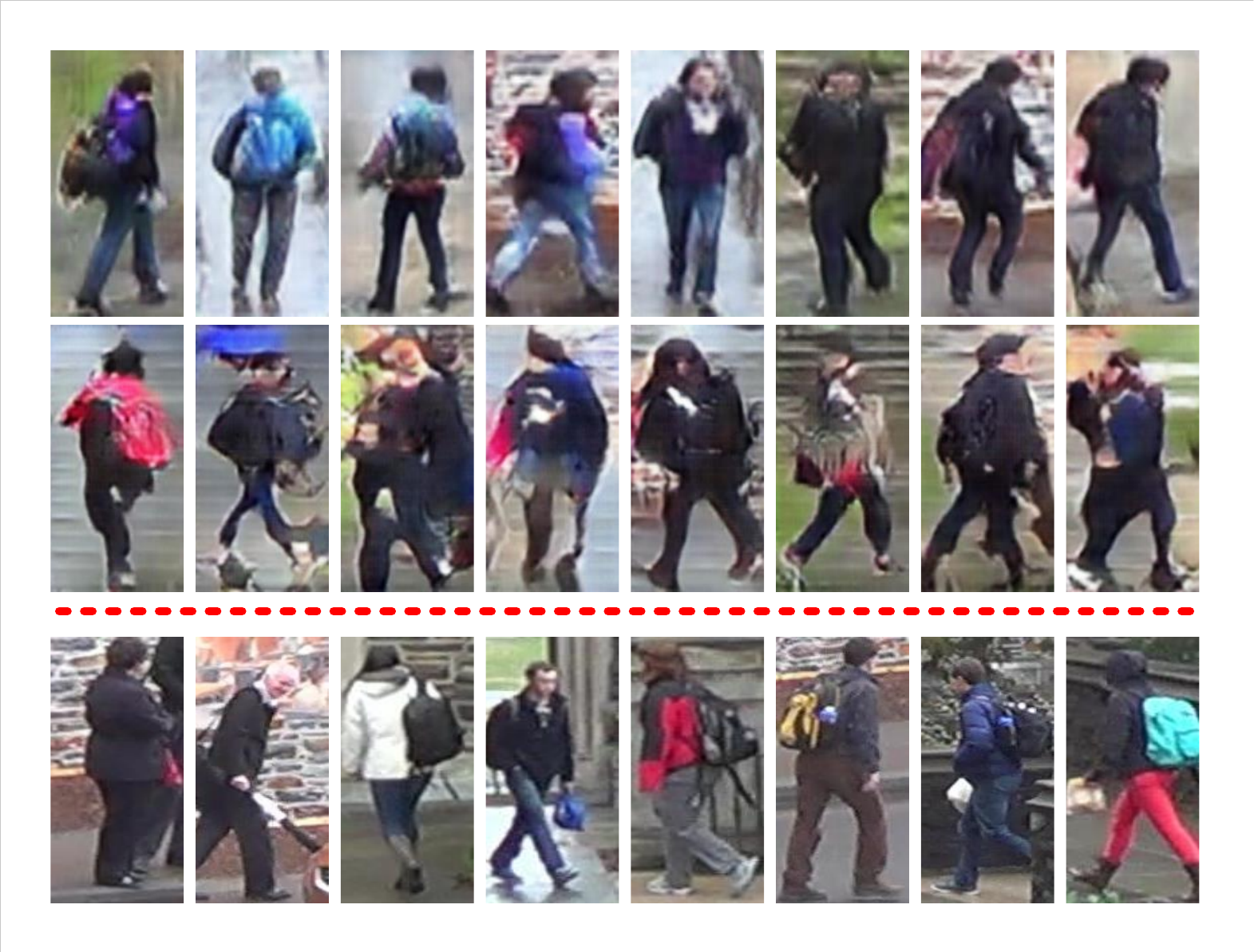}
        \label{fig:duke_wgangp}
    }
    \subfigure[VIPeR.]{
        \includegraphics[width=0.6\columnwidth]{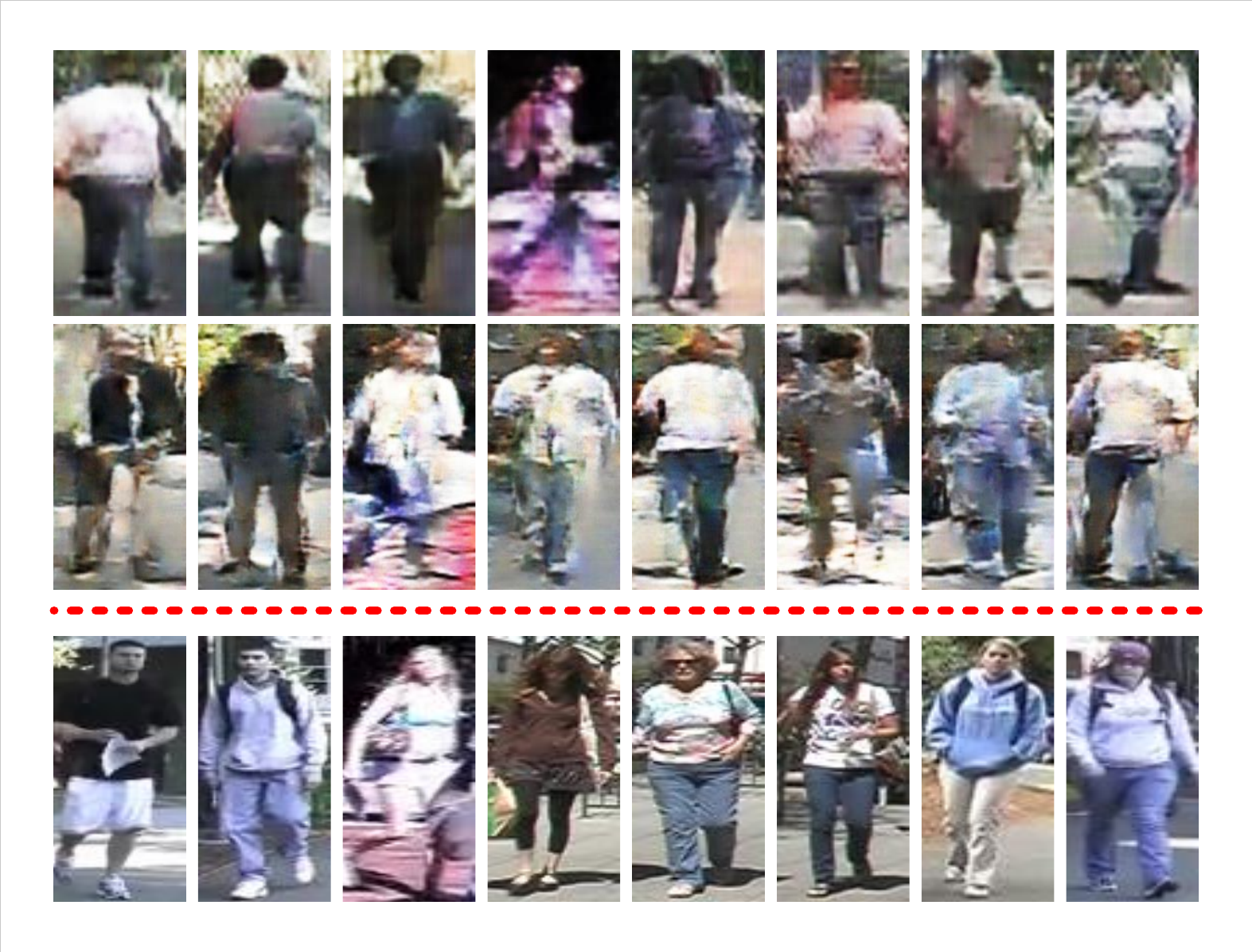}
        \label{fig:viper_wgangp}
    }
    \caption{Examples of generated and real person images. (a)-(c) show the generated person images (first two rows) and real person images (the third row) on Market-1501, DukeMTMC-reID, and VIPeR respectively. Images in the first and second rows are respectively generated by the WGAN-GP \cite{gulrajani2017improved} and the DCGAN \cite{radford2015unsupervised}.}\label{fig:wgangp_images}
\end{figure*}

\begin{table}[t]
  \centering
  \caption{Comparison between using generated data by DCGAN and WGAN-GP. Two approaches are used, including LSRO and the proposed dMpRL-II. Experiments conducted on three datasets: Market-1501, DukeMTMC-reID, and VIPeR. Rank-1 accuracy and mAP are listed.}
    \begin{tabular}{c|c|c|c|c}
    \toprule[1.1pt]\hline
    \multirow{3}{*}{Method} & \multicolumn{4}{c}{Market-1501} \\ \cline{2-5}   
    & \multicolumn{2}{c|}{DCGAN \cite{radford2015unsupervised}} & \multicolumn{2}{c}{WGAN-GP \cite{gulrajani2017improved}} \\ \cline{2-5}
	& mAP & rank-1 & mAP  & rank-1  \\
    \hline
    LSRO \cite{ZhengZY17} & 56.33\%	& 78.21\% & 55.53\%	& 78.32\% \\
    dMpRL-II & 58.59\% & 80.37\% & 59.04\% & 79.75\% \\ \hline
    & \multicolumn{4}{c}{DukeMTMC-reID} \\ \hline
    LSRO \cite{ZhengZY17} & 46.66\% & 66.92\% & 46.79\%	& 66.97\% \\
    dMpRL-II & 48.58\% & 68.24\% & 49.30\% & 68.76\% \\ \hline
    & \multicolumn{4}{c}{VIPeR} \\ \hline
    LSRO \cite{ZhengZY17} & 49.41\%	& 43.39\% & 48.47\%	& 43.14\% \\
    dMpRL-II & 52.25\% & 46.60\% & 52.16\% & 46.39\% \\
    \hline\bottomrule[1.1pt]
    \end{tabular}%
  \label{tab:wgangp-result}%
\end{table}

\begin{table*}[t]
  \centering
  \caption{Comparison of our results with the published state-of-the-art methods. The best and the second-best results are shown in \textbf{bold} and \underline{underline}, respectively. Rank-1 accuracy and mAP are listed. The ReK means re-ranking.}
    \begin{tabular}{l|l|c|c|c|c|c|c|c|c|c|c}
    \toprule[1.1pt]\hline
    \multicolumn{2}{c|}{\multirow{4}{*}{Method}} &\multicolumn{8}{c|}{Large-Scale Datasets} &\multicolumn{2}{c}{Small-Scale Datasets} \\ \cline{3-12}
    \multicolumn{2}{c|}{} & \multicolumn{4}{c|}{Market-1501} & \multicolumn{2}{c|}{DukeMTMC-reID} & \multicolumn{2}{c|}{CUHK03} & VIPeR & CUHK01\\ \cline{3-12}  
    \multicolumn{2}{c|}{} & \multicolumn{2}{c|}{Single Query} & \multicolumn{2}{c|}{Multiple Query} & \multicolumn{2}{c|}{Single Query} & \multicolumn{2}{c|}{Single Query (detected)} & Single Query & Multiple Query\\ \cline{3-12}  
    \multicolumn{2}{c|}{} & mAP   & rank-1 & mAP   & rank-1 & mAP   & rank-1 & mAP  & rank-1  & rank-1 & rank-1\\ \hline
    Gate-reID \cite{varior2016gated} & ECCV16  & 39.55\%  & 65.88\% & 48.45\%  & 76.04\%  & --  &--  & 58.84\%  & 68.10\% &37.80\% &--\\
    SI-CI \cite{wang2016joint} & CVPR16  & --  & -- & --  & --  & --  &--  & --  & 52.17\% &35.76\% &--\\
    GOG+XQDA \cite{matsukawa2016hierarchical} & CVPR16  & --  & -- & --  & --  & --  &--  & --  & 65.50\% &49.70\% &57.80\%\\
    SCSP \cite{chen2016similarity} & CVPR16  & 26.35\%  & 51.90\% & --  & --  & --  &--  & --  & -- &53.54\% &--\\
    DNS \cite{zhang2016learning}& CVPR16 & 35.68\%  & 61.02\% & 46.03\%  & 71.56\%  & --  &--  & --  & 54.70\% &51.17\% &69.09\%\\
    Resnet+OIM \cite{xiao2017joint}& CVPR17 & -- & 82.10\% & --  & -- & -- & 68.10\% & --  & -- &-- &-- \\
    Latent Parts \cite{li2017learning}& CVPR17  & 57.53\% & 80.31\% & 66.70\% & 86.79\% & -- & -- & -- & 67.99\% &38.08\% &-- \\
    P2S \cite{zhou2017point}& CVPR17  & 44.27\% & 70.72\% & 55.73\% & 85.78\% & --     & --     & --     & -- &-- &-- \\
    ReRank \cite{zhong2017re}& CVPR17 & 63.63\% & 77.11\% & --     & --     & --     & --     & --     & --  &-- &-- \\
    CADL \cite{lin2017consistent}& CVPR17 & 55.60\% & 80.90\% & --   & --     & --     & --     & --     & -- &-- &-- \\
    SpindleNet \cite{zhao2017spindle}& CVPR17 & --   & 76.90\% & --     & --     & --     & --     & --     & -- &\underline{53.80\%} &\textbf{79.90\%} \\
    SSM \cite{bai2017scalable}& CVPR17 & 68.80\% & 82.21\% & 76.18\% & 88.18\% & --   & --  & --  & 72.70\% &53.73\% &-- \\
    JLML \cite{li2017person}& IJCAI17 & 65.50\% & 85.10\% & 74.50\% & 89.70\% & --  & --  & --   & 80.60\% &50.20\% &76.70\% \\
    SVDNet \cite{sun2017svdnet}& ICCV17 & 62.10\% & 82.30\% & --  & -- & 56.80\% & 76.70\% & 84.80\% & 81.80\% &-- &-- \\
    PDC \cite{su2017pose}& ICCV17 & 63.41\% & 84.14\% & --   & --   & --  & --  & --  & 78.29\%  &51.27\%  &-- \\
    Part Aligned \cite{zhao2017deeply}& ICCV17  & 63.40\% & 81.00\% & --  & --   & --  & --  & --  & 81.60\%  &48.70\%  &75.00\% \\
    LSRO \cite{ZhengZY17}& ICCV17& 66.07\% & 83.97\% & 76.10\% & 88.42\% & 47.13\% & 67.68\% & 87.40\% & 84.60\% &-- &-- \\ \hline
    \multicolumn{2}{l|}{Identif \cite{zheng2016person, zheng2016person2}} & 52.68\% & 74.08\% & 64.95\% & 82.06\% & 42.20\% & 61.94\% & 68.36\% & 63.10\% & 40.76\% & 65.33\%\\
    \multicolumn{2}{l|}{\textbf{Identif+dMpRL-II}} & 58.59\% & 80.37\% & 70.22\% & 86.47\% & 48.58\% & 68.24\% & 73.48\% & 68.68\% & 46.60\% & 68.81\% \\ \hline
    \multicolumn{2}{l|}{Two-stream \cite{zheng2016discriminatively}} & 64.09\% & 81.83\% & 73.65\% & 86.82\%     & 51.40\% & 72.62\% & 85.20\% & 81.88\% & 51.84\% & 77.78\% \\
    \multicolumn{2}{l|}{\textbf{Two-stream+dMpRL-II}} & \underline{67.53\%} & \underline{85.75\%} & \underline{77.85\%} & \underline{89.88\%} & \underline{58.56\%} & \underline{76.81\%} & \underline{87.53\%} & \underline{85.42\%} & \textbf{54.65\%} & \underline{78.83\%} \\
    \multicolumn{2}{l|}{\textbf{Two-stream+dMpRL-II+ReK}} & \textbf{81.18\%} & \textbf{87.96\%} & \textbf{86.53\%} & \textbf{90.97\%} & \textbf{74.54\%} & \textbf{81.28\%} & \textbf{90.16\%} & \textbf{88.00\%} & 53.22\% & 78.08\% \\
	\hline\bottomrule[1.1pt]
    \end{tabular}%
  \label{tab:state_of_the_art}%
\end{table*}%

\subsection{Comparison with Different GAN Models}
\label{sec:Comparison_with_Different_Generation_Models}
In addition to the DCGAN, other GAN models such as WGAN-GP \cite{gulrajani2017improved} has demonstrated its superior in generating high quality person images. We attempt to generate data using the WGAN-GP. Then, the relationship between the quality of generated images and our proposed MpRL can be testified by using different GAN models. In this experiment, two large and one small-scale datasets are used individually to generate images. Figure \ref{fig:wgangp_images} shows the generated data by using different GAN models. It can be observed that the WGAN-GP exhibits better capability of generating person images on these datasets. In order to verify the impacts of image quality created by different GAN approaches, we compare the performance of the proposed MpRL on the two different generated data sets. Table \ref{tab:wgangp-result} lists the comparison results. It is observed that by using generated data with different quality through different GAN approaches, the re-ID performance is not significantly affected. This is because these generated data are employed to improve the performance of CNN models by its regularization power instead of providing more actual subjects beyond the scope of the raw dataset in training. Therefore, better generated data can bring superior perceptual quality but cannot dramatically boost the effectiveness of regularizer although some marginal improvements can be observed.

\subsection{Comparison with The State-of-the-art Methods}
\label{sec:Comparison_with_the_state-of-the-art_methods}
Although the main contribution in this paper focuses on using the generated data to improve the performance of CNNs, but not on producing a state-of-the-art result, we still compare our result with several state-of-the-art methods. Table \ref{tab:state_of_the_art} lists the comparison results. It is clear to see that although the performance of the original Two-stream network is competitive, it still be inferior to many methods such as Resnet+OIM \cite{xiao2017joint}, SSM \cite{bai2017scalable}, JLML \cite{li2017person}, SVDNet \cite{sun2017svdnet}, and PDC \cite{su2017pose}. However, by incorporating with the proposed dMpRL-II, the Two-stream network achieves the state of the art compared with other methods on Market-1501, DukeMTMC-reID, CUHK03 and VIPeR. To achieve better performance, after obtaining the rank list by sorting the similarity of gallery images to a query, a re-ranking method \cite{zhong2017re} is adopted to further boost our performance. The combination of the dMpRL-II and re-ranking on the Two-stream network achieves the best results on the three large-scale datasets. However, the re-ranking approach cannot further improve the performance of the two small-scale datasets with limited number of testing person identities. We find that the rank-1 accuracy of the DPFL method \cite{chen2017person} proposed in the ICCV17 workshop is slightly higher than our result on Market-1501 (88.90\% in single query and 92.30\% in multiple query). However, DPFL uses an ensemble deep model with multiple granularity inputs for each image. Our Two-stream network just utilizes a single model and outperforms the DPFL on CUHK03 by a large margin in mAP even without re-ranking (mAP: 87.53\% (our) vs. 78.10\% (DPFL), rank-1: 85.42\% (our) vs. 82.00\% (DPFL)). Also, the performance of the Spindle \cite{zhao2017spindle} approach is slightly higher than ours on CUHK01 (79.90\% \textit{vs.} 78.83\%). Since VIPeR and CUHK01 are two small-scale datasets, nine different person re-ID datasets are used to pre-train the SpindleNet model and then fine-tuning on the two small datasets respectively. We also use the fine-tuning strategy on these two datasets, but only three datasets are involved in the pre-training stage (see \ref{sec:CNN}). Except for the CUHK01 dataset, our performance outperforms the SpindleNet on the other small-scale dataset VIPeR and the three large-scale re-ID datasets simultaneously.

\section{Conclusion}
\label{sec:conclusion}
In this paper, we propose a new virtual label MpRL for the generated data by GAN. To train a CNN, MpRL is used as virtual label assigned to generated data. These data are used for semi-supervised learning. Two CNNs are adopted to show the effectiveness of the proposed MpRL. Experiments demonstrate that generated data can effectively improve the performance of the two CNNs trained with the proposed MpRL. Compared with the previous state-of-the-art method LSRO \cite{ZhengZY17}, MpRL can always achieve better improvements. In the future, considering the capability of GAN, we will continue to investigate virtual labels used on generated data for semi-supervised learning and apply the proposed method to other fields.


\ifCLASSOPTIONcaptionsoff
  \newpage
\fi


\bibliographystyle{abbrv}
\bibliography{mybibfile}



\end{document}